\documentclass{article}

\usepackage{microtype}
\usepackage{graphicx}
\usepackage{booktabs} %

\usepackage{algorithm,algpseudocode}

\usepackage[misc]{ifsym}
\usepackage{fontawesome5}

\usepackage{hyperref}

\usepackage{multirow}

\usepackage[accepted]{icml2025}

\usepackage[font=small,labelfont=bf]{caption}
\usepackage{subcaption}

\usepackage{amsmath}
\usepackage{amssymb}
\usepackage{mathtools}
\usepackage{amsthm}

\usepackage[frozencache,cachedir=.]{minted}

\usepackage{xcolor}

\usepackage[capitalize,noabbrev]{cleveref}

\theoremstyle{plain}

\theoremstyle{definition}

\theoremstyle{remark}

\usepackage[textsize=tiny]{todonotes}

\usepackage{pgfplots}
\usetikzlibrary{pgfplots.groupplots}
\pgfplotsset{compat=1.3}
\usepackage{tikz}
\usetikzlibrary{patterns}

\usepackage{microtype}
\usepackage{inconsolata}

\usepackage[font=small,labelfont=bf]{caption}
\definecolor{mediumpurple}{rgb}{0.58, 0.44, 0.86}
\usepackage[most]{tcolorbox}

\usepackage{tcolorbox}

\definecolor{battleshipgrey}{rgb}{0.3, 0.3, 0.3}
\definecolor{brilliantrose}{rgb}{1.0, 0.33, 0.64}
\definecolor{americanrose}{rgb}{1.0, 0.01, 0.24}
\definecolor{jweigreen}{rgb}{0,0.45,0.24}
\definecolor{bluegray}{rgb}{0.1, 0.1, 0.4}
\definecolor{ao(english)}{rgb}{0.0, 0.5, 0.0}
\definecolor{blanchedalmond}{rgb}{1.0, 0.92, 0.8}
\definecolor{atomictangerine}{rgb}{1.0, 0.6, 0.4}
\definecolor{chocolate(web)}{rgb}{0.82, 0.41, 0.12}
\definecolor{bananayellow}{rgb}{1.0, 0.88, 0.21}
\definecolor{goldenbrown}{rgb}{0.6, 0.4, 0.08}
\definecolor{aliceblue}{rgb}{0.94, 0.97, 1.0}
\definecolor{beige}{rgb}{0.96, 0.96, 0.86}
\definecolor{babyblue}{rgb}{0.54, 0.81, 0.94}
\definecolor{camel}{rgb}{0.76, 0.6, 0.42}
\definecolor{cinnamon}{rgb}{0.82, 0.41, 0.12}
\definecolor{deepskyblue}{rgb}{0.0, 0.75, 1.0}
\definecolor{frenchblue}{rgb}{0.0, 0.45, 0.73}
\definecolor{classicrose}{rgb}{0.98, 0.8, 0.91}
\definecolor{frenchrose}{rgb}{0.96, 0.29, 0.54}
\definecolor{frenchlilac}{rgb}{0.53, 0.38, 0.56}
\definecolor{frenchbeige}{rgb}{0.65, 0.48, 0.36}
\definecolor{applegreen}{rgb}{0.55, 0.71, 0.0}
\definecolor{dartmouthgreen}{rgb}{0.05, 0.5, 0.06}
\definecolor{turquoisegreen}{rgb}{0.63, 0.84, 0.71}
\definecolor{darkseagreen}{rgb}{0.56, 0.74, 0.56}
\definecolor{columbiablue}{rgb}{0.61, 0.87, 1.0}
\definecolor{rufous}{rgb}{0.66, 0.11, 0.03}
\definecolor{cyan(process)}{rgb}{0.0, 0.72, 0.92}
\definecolor{crimsonglory}{rgb}{0.75, 0.0, 0.2}
\definecolor{yaleblue}{rgb}{0.06, 0.3, 0.57}

\usepackage{tcolorbox}

\usepackage[symbol]{footmisc} %

\renewcommand{\thefootnote}{\fnsymbol{footnote}} %

\icmltitlerunning{Dynamic Cheatsheet: Test-Time Learning with Adaptive Memory}

\begin{document}

\onecolumn
\icmltitle{Dynamic Cheatsheet: Test-Time Learning with Adaptive Memory}

\begin{icmlauthorlist}
\icmlauthor{Mirac Suzgun}{stan}
\icmlauthor{Mert Yuksekgonul}{stan}
\icmlauthor{Federico Bianchi}{togetherai}
\icmlauthor{Dan Jurafsky}{stan}
\icmlauthor{James Zou}{stan,togetherai}
\end{icmlauthorlist}

\icmlaffiliation{stan}{Stanford University}
\icmlaffiliation{togetherai}{Together AI}

\icmlcorrespondingauthor{Mirac Suzgun}{msuzgun@stanford.edu}

\icmlkeywords{Machine Learning, ICML}

\vskip 0.3in

\printAffiliationsAndNotice{}  %

\begin{abstract}
Despite their impressive performance on complex tasks, current language models (LMs) typically operate in a vacuum: Each input query is processed separately, without retaining insights from previous attempts. Here, we present \emph{Dynamic Cheatsheet} (DC), a lightweight framework that endows a black-box LM with a persistent, evolving memory. Rather than repeatedly re-discovering or re-committing the same solutions and mistakes, DC enables models to store and reuse accumulated strategies, code snippets, and general problem-solving insights at inference time. This test-time learning enhances performance substantially across a range of tasks without needing explicit ground-truth labels or human feedback. Leveraging DC, Claude 3.5 Sonnet’s accuracy more than doubled on AIME math exams once it began retaining algebraic insights across questions. Similarly, GPT-4o’s success rate on the Game of 24 puzzle increased from about 10\% to 99\% after the model discovered and reused a Python-based solution. In tasks prone to arithmetic mistakes, such as balancing equations, DC enabled GPT-4o and Claude to reach near-perfect accuracy by recalling previously validated code, whereas their baselines stagnated around 50\%.  Beyond arithmetic challenges, DC yields notable accuracy gains on knowledge-demanding tasks. Claude achieved a 9\% improvement in GPQA-Diamond and an 8\% boost on MMLU-Pro Engineering and Physics problems. Crucially, DC’s memory is self-curated, focusing on concise, transferable snippets rather than entire transcripts, thereby facilitating meta-learning and avoiding context ballooning. Unlike fine-tuning or static retrieval methods, DC adapts LMs’ problem-solving skills on the fly, without modifying their underlying parameters, and offers a practical approach for continuously refining responses and cutting routine errors. Overall, our findings present DC as a promising approach for augmenting LMs with persistent memory, bridging the divide between isolated inference events and the cumulative, experience-driven learning characteristic of human cognition.\footnote{{\faGithub}  We release all our data, results, and code at \url{http://github.com/suzgunmirac/dynamic-cheatsheet}.}
\begin{figure*}[h]
\centering
\includegraphics[width=0.94\textwidth]{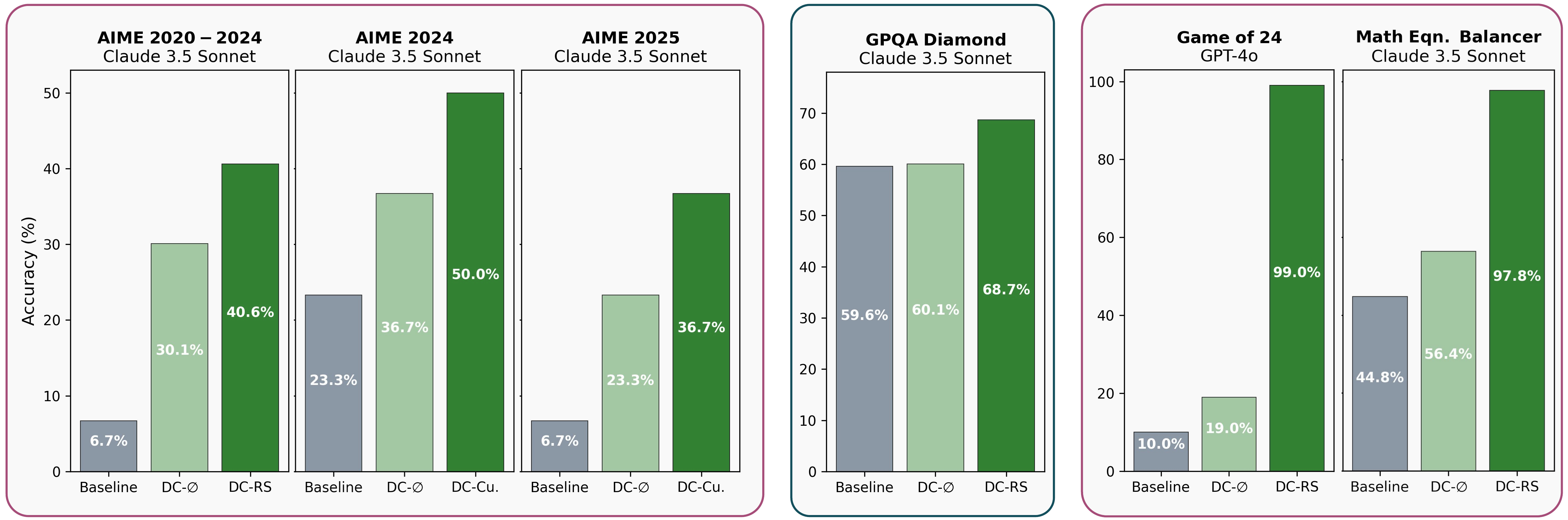}
\caption{
Comparison of different baselines and Dynamic Cheatsheet (DC) variants on challenging reasoning benchmarks, including AIME exams and GPQA-Diamond. Baseline represents a standard prompting approach with minimal guidance, while DC-$\emptyset$ (a stronger baseline) contains explicit structured instructions for problem solving, as well as for Python code generation and execution, but lacks a memory component. Our proposed DC-Cu and DC-RS variants incorporate an evolving, text-based memory to enhance inference-time learning. Results (accuracy, \%) demonstrate substantial improvements, with Claude 3.5 Sonnet gaining 27\% on AIME 2024 and 30\% on AIME 2025 under DC-Cu. In Game of 24, GPT-4o leaps from 10\% (baseline) to 99\% under DC-RS, reflecting its ability to retain and apply Python-based solutions efficiently. Similarly, Claude 3.5 Sonnet’s accuracy more than doubles in Math Equation Solver, reaching 98\%. Overall, these findings highlight the impact of test-time learning through controlled memory augmentation and efficient retrieval.}
\label{fig:splashfigure}
\end{figure*}

\end{abstract}

\renewcommand{\thefootnote}{\arabic{footnote}}
\setcounter{footnote}{0}

\newpage

\twocolumn

\section{Introduction}
\label{sec:introduction}

Modern large language models (LLMs) can tackle complex reasoning tasks, answer various questions, and generate extensive texts. Yet they still suffer from one critical limitation: once deployed, these models are fixed prior to deployment and typically retain no explicit or implicit memory of past questions, successes, or mistakes during inference. They approach each new problem \emph{de novo}, often re-deriving the same insights—and re-committing the same errors. In contrast, human cognition stands on a foundation of incremental learning, continuously internalizing new experiences and solutions into a persistent mental model. 

In this work, we present \emph{Dynamic Cheatsheet} (DC), a simple and intuitive framework that endows black-box LLMs with a persistent, evolving memory at inference time. Rather than fine-tuning weights (for instance, through dynamic evaluation~\citep{krause2019dynamic} or domain adaptation~\citep{gururangan2020don}) or retrieving facts from a massive \emph{static} corpus (as in traditional retrieval-augmented generation systems~\citep{guu2020retrieval,zhang2024raft}), DC dynamically curates a compact library of reusable strategies, solution sketches, and code snippets. Either before or after each query, DC enables the system to decide which lessons to store, what to discard, and how to refine existing entries—thus effectively “learning” from successes and failures. It is a flexible online-learning approach that enables a black-box LLM to improve itself without needing any explicit ground truth labels or human feedback.

\begin{figure}[ht]
\centering
\includegraphics[width=0.99\linewidth]{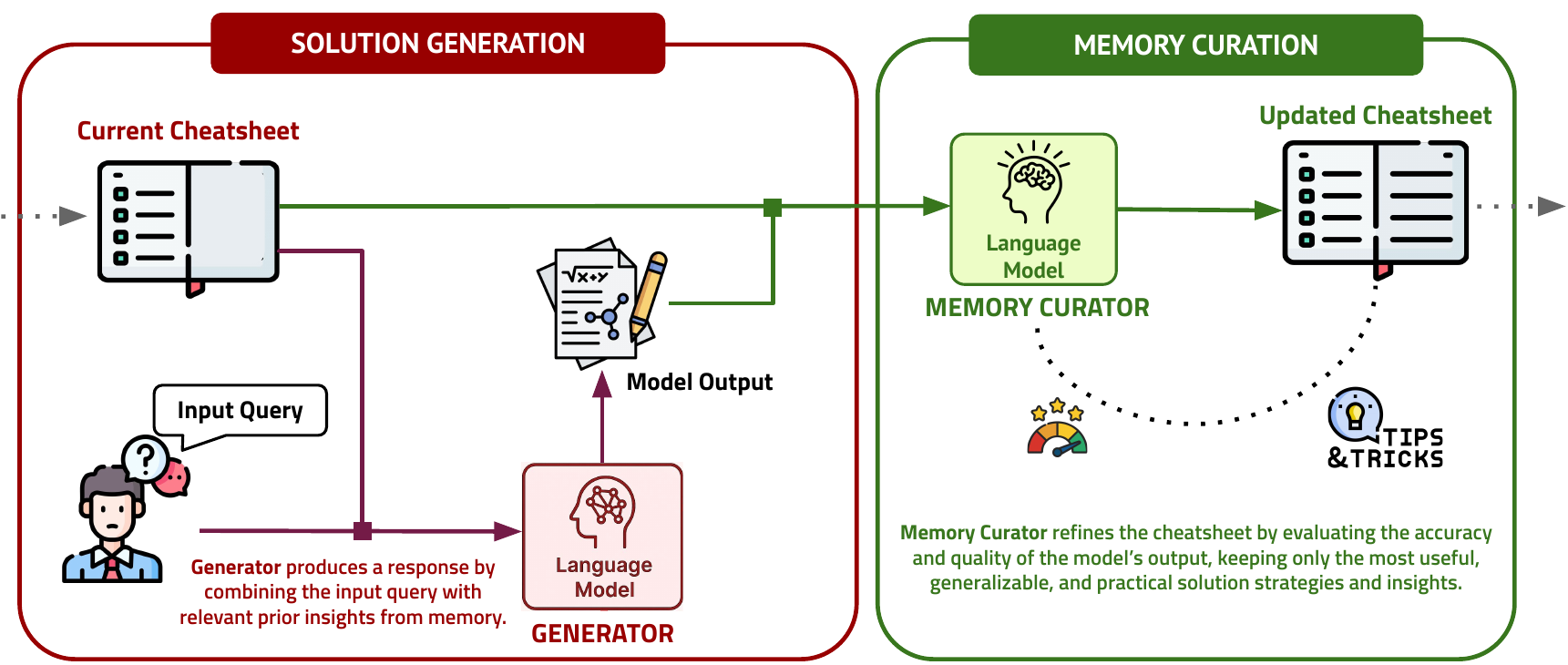}
\vspace{-0.2em}
\caption{
Illustration of \emph{Dynamic Cheatsheet} (DC-Cu variant).
}
\vspace{0.4em}
\label{fig:dc_illustration}
\end{figure}

The overall workflow of DC is intuitive and compelling. In one version of DC (DC-Cu.), when presented with a new query, the LM first consults its external memory to see if any prior insights, strategies or relevant model solutions have been stored. It then proposes a solution by combining the retrieved insights with its own internal reasoning capabilities. Upon generating an answer, it then proceeds to a curation phase that updates the memory: If the approach seems to be correct, useful, or practical, DC codifies it in its memory for future use; if an error surfaces, DC may revise or prune faulty heuristics. This all happens without gradient-based parameter updates, so computational overhead remains modest, and compatibility with black-box APIs (e.g., GPT-4 or Claude) is fully preserved. \emph{See} Figure~\ref{fig:dc_illustration}.

We tested DC across multiple challenging benchmarks and observed that it increases performance and reduces repetitive mistakes. On AIME 2024, Claude 3.5 Sonnet jumped from 23\% to 50\% accuracy, more than doubling its baseline score, by retaining algebraic and combinatorial insights. Likewise, it gained 30\% accuracy on AIME 2025. Notably, these improvements hold in  knowledge-intensive tasks as well. On GPQA-Diamond, which tests specialized domain questions, DC lifted Claude by over 9\%. In MMLU-Pro Engineering and Physics, it provided up to an 8\% boost in performance by allowing the model to maintain a ``toolkit'' of formulas and general problem-solving patterns.

\begin{figure}[!t]
\centering
\includegraphics[width=0.99\linewidth]{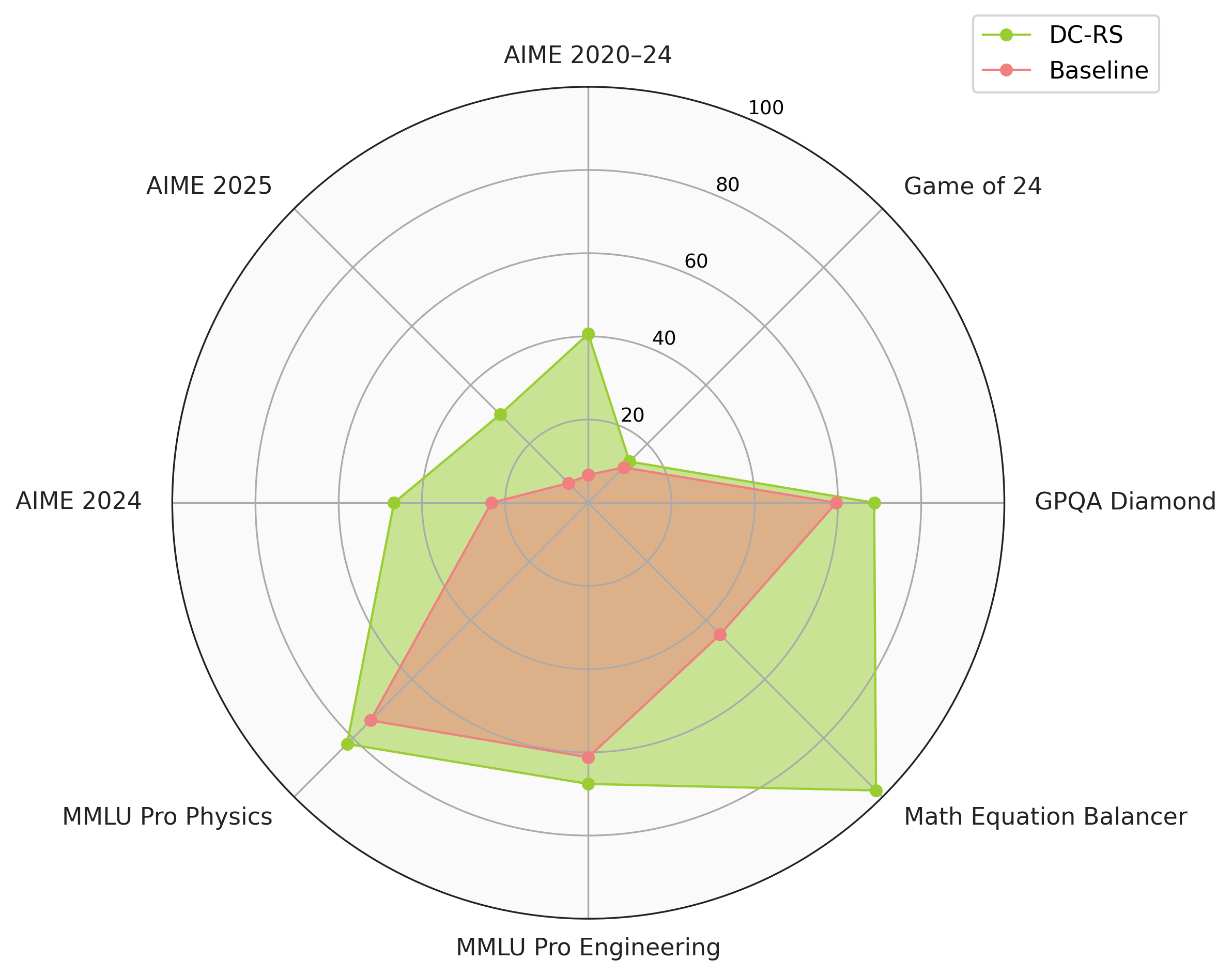}
\vspace{-0.2em}
\caption{
Overall task performance of Claude 3.5 Sonnet under the baseline prompting approach with minimal instructions (BL) and Dynamic Cheatsheet with Retrieval \& Synthesis (DC-RS). }
\label{fig:task-performance}
\end{figure}

An even more striking and compelling example is the Game of 24, a puzzle that requires the solver to combine four digits into an arithmetic expression equaling 24. GPT-4o’s baseline performance (10\%) increased to 99\% under DC. Early in the test sequence, the model discovered that an efficient Python brute-force solver eliminated all manual guesswork. Once this snippet was stored, GPT-4o simply retrieved it for subsequent queries, avoiding manual arithmetic entirely. We saw a similar pattern in Math Equation Balancer, where GPT-4o and Claude soared from 45-50\% to 98–100\% by ``recalling'' a straightforward code-based approach instead of manually fumbling with numeric manipulations.

Nonetheless, DC is not a panacea. We found that smaller models, such as GPT-4o-mini, benefit from DC in limited amounts. These models  generate too few correct  solutions in these challenging tasks in the first place, leaving the memory populated with flawed or incomplete strategies. Worse, they struggle to refine stored content. 
DC can amplify the strengths of models that can already produce high-quality outputs, but not fix foundational gaps in reasoning.

We also note that DC differs from naive ``append the entire conversation history'' in-context learning approaches. Under DC, memory is carefully curated, focusing on succinct, useful, and transferable knowledge over raw transcripts. This prevents ballooning context lengths~\citep{liu2024lost} and helps ensure that repeated retrieval remains tractable. Indeed, part of DC’s contribution is in formalizing a mechanism for selective, evolving retention—storing just enough to solve the next set of tasks without drowning in an ever-growing text buffer. \emph{Cf.}~\citep{karpicke2008critical,roediger2011critical,karpicke2011retrieval}

\begin{figure*}[!t]
\centering
\includegraphics[width=1\linewidth]{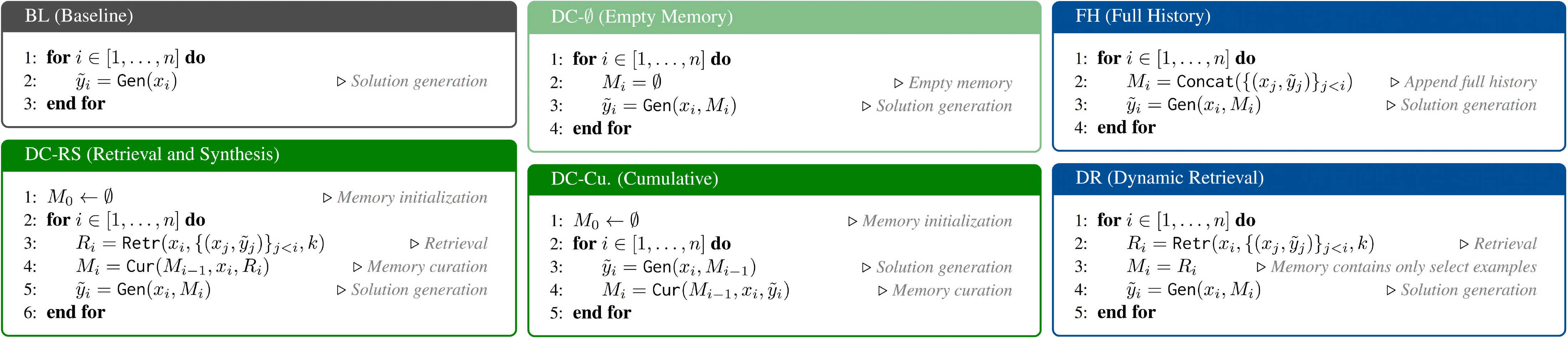}
\caption{Algorithmic illustration of the Dynamic Cheatsheet (DC)-based approaches and other baseline methods. Here,  $\texttt{Gen}$ represents the solution generator model,  $\texttt{Cur}$ the memory curator, and  $\texttt{Retr}$ the retriever. While we use the same black-box LLMs for both generation and curation, we differentiate their roles via task-agnostic instructions (prompts). The retrieval mechanism ranks historical inputs based on cosine similarity with the current query, selecting the most relevant past examples along with their generated solutions.}
\label{fig:pseudocodes}
\end{figure*}

\section{Dynamic Cheatsheet (DC) Methodology}
\label{sec:methodology}

DC, in its core, includes an external, non-parametric  memory that evolves in tandem with the LLM’s inference process. Rather than fine-tuning the underlying weights, DC tracks successes and failures of the model at test time, then selectively stores heuristics, strategies, or short textual artifacts that can guide the LLM in future instances. Notably, this approach respects the black-box nature of many commercial LLM APIs: no gradient-based updates are required, and the model’s core parameters remain untouched.

\subsection{DC: Building Blocks and Iterative Loop}

The DC framework consists of two core modules: \emph{generation} and \emph{curation}. Both modules can easily operate on top of the same LM (prompted differently) or on separate LMs. 

\subsubsection{Solution Generation with Memory}
\label{sec:methodology:generation}

Let's consider a sequence of inputs $(x_1, x_2, \dots, x_n)$,  where each $x_i \sim \mathcal{D}_{\text{test}}$ indicates a new query or problem posed to the model sampled from the same distribution $\mathcal{D}_{\text{test}}$~(a typical setting in online learning). The distribution $\mathcal{D}_{\text{test}}$ is unknown to us. At the $i$-th step, the model is provided with both the new query $x_i$ and the current memory state $M_i$, which captures knowledge gleaned from previous successes and failures. We denote the solution generator by $\texttt{Gen}$:
\begin{equation}
\tilde{y}_i = \texttt{Gen}(x_i, M_i)
\label{eq:solution-generation}
\end{equation}
\noindent Here, $\tilde{y}_i$ is the candidate solution produced by the model. $M_i$ helps condition the model to reuse or adapt previously stored solutions, insights, techniques, or heuristics. %

\subsubsection{Memory Curation Step}
\label{sec:methodology:curation}

After the generator produces its answer $\tilde{y}_i$ to $x_i$, the curator, $\texttt{Cur}$, updates the current content of the memory:
\begin{equation}
M_{i+1} = \texttt{Cur}(M_i, x_i, \tilde{y}_i)
\label{eq:memory-curation}
\end{equation}

During memory curation, $\texttt{Cur}$ mainly considers: (i) \emph{the usefulness and generalizability of the newly produced answer} (i.e., if $\tilde{y}_i$ is correct or provides valuable and generalizable insights, it is distilled into a form suitable for later reference), (ii) \emph{refinement or removal of existing memory entries} (i.e., if an existing memory entry was incorrect or superseded by a more efficient or versatile strategy, $\texttt{Cur}$ may remove or update it), and (iii) \emph{clarity and compactness of the entire memory} (i.e., memory entries are consolidated to retain succinct, high-impact references and heuristics).

$\texttt{Cur}$ does not have access to ground-truth labels; so, it has to assess the correctness and efficiency of the solutions by itself before updating the memory. In our experiments, we instruct a single model to perform this crucial step. Yet, in practice, $\texttt{Cur}$ can be implemented as a series of steps that instruct multiple tools and models, through different prompts, to verify the validity and efficiency of the solution and to transform the raw solution text into even more generalizable, reliable, and efficient strategies, insights, and code snippets.

We refer to this version of DC above as \textbf{DC-Cu} (short for DC-Cumulative). Under DC-Cu, the system first performs solution generation based on the current memory (Eqn.~\ref{eq:solution-generation}) and then updates the memory (Eqn.~\ref{eq:memory-curation}), by cumulatively expanding and refining the memory items thus far. Unlike DC-RS, which is discussed in the next part, DC-Cu, does not contain a retrieval component, however.

\subsection{DC with Retrieval \& Synthesis (DC-RS)}

DC-Cu has two potential drawbacks. \emph{First}, it updates the memory after processing an input query, rather than refining it before generating a response. This means the model lacks the opportunity to incorporate new insights from the current query while reasoning through its solution. \emph{Second}, DC-Cu does not store or revisit past input-output pairs unless explicitly retained in memory. This omission prevents the model from directly retrieving and leveraging historical responses, which can be particularly valuable in benchmarks covering diverse topics or domains (e.g., GPQA-Diamond).

To address these issues, \textbf{DC-RS} modifies the sequence of memory updates and introduces a retrieval mechanism, $\texttt{Retr}$, into the curation process. $\texttt{Retr}$ allows the model to retrieve the most relevant past input-output pairs from its knowledge base. By refining the memory before responding and retrieving prior cases when needed, DC-RS enhances the model’s adaptability and reasoning efficiency.

DC-RS first retrieves\footnote{We used OpenAI's \texttt{text-embedding-3-small} model to map input queries (raw questions) to embedding vectors.} top-$k$\ most similar inputs, along with their model-generated outputs, from previously seen examples, which we denote by $R^{(k)}_i$ (or simply $R_i$).\footnote{We set $k$ to 3 in  all our experiments. (Initially, we considered  higher top-$k$ values such as 5 and 7, but the gain was insignificant.)} It then passes these select examples, $R_i$, along with the most recent memory content, $M_{i-1}$, to the curator to update the memory, that is to get $M_i$. Finally, it uses the generator to produce $\tilde{y}_i$, given $x_i$ and $M_i$. We summarize all these steps below:
\begin{align}
    R_i &= \texttt{Retr}(x_i, \{(x_j, \tilde{y}_j)\}_{j < i}, k) \\
    M_i &= \texttt{Cur}(M_{i-1}, x_i, R_i) \\
    \tilde{y}_i &= \texttt{Gen}(x_i, M_i)
\end{align}

\subsection{Baselines}

To quantify the efficacy of memory-driven test-time learning, we compare DC and its variants to four baselines:

\textbf{(1) Baseline prompting (BL).} This plain ``vanilla'' prompting approach, with minimal instructions, simply prompts the model without any iterative memory or retrieval mechanism. It reflects traditional one-off inference.\footnote{Please refer to Figure~\ref{fig:baseline-prompt} to see the full instruction (prompt) used in BLh. We experimented with the zero-shot CoT approach~\citep{kojima2022large} in our preliminary experiments, but it did not yield any gains~\citep{arcuschin2025chainofthought}. We, therefore, did not include it as a baseline method in our experiments.}

\textbf{(2) DC-$\emptyset$ (empty memory).} To isolate the effect of memory curation, this DC baseline always keeps the memory content effectively empty.\footnote{We adopt the generator prompt template used in DC-RS, namely  Figure~\ref{fig:generator-prompt}, for DC-$\emptyset$, though we replace the memory placeholder with the text ``(empty cheatsheet)''.}  DC-$\emptyset$ allows us to measure how much performance improvement arises purely from storing and reusing knowledge over time. While there is no continuous knowledge storage or strategy reuse, this method follows the instructions in Figure~\ref{fig:generator-prompt} and is therefore a strong baseline.

\textbf{(3) Full-History Appending (FH).} This is a naive approach that appends the entire conversation history to the model input without any curation or truncation.\footnote{We consider and test this baseline only on AIME 2024 and AIME 2025, which are relatively small in their size (each contains 30 examples) compared to other benchmarks.} FH can exceed context-window limits and include redundant or low-value information, but nonetheless, it provides a useful comparison for methods that actively curate content.\footnote{We use the generator prompt template in Figure~\ref{fig:generator-prompt} again, but include the entire raw input-output pairs from the previous steps in the memory---without any curation, truncation, or synthesis.}

\textbf{(4) Dynamic Retrieval (DR).} A final baseline uses retrieval but no curation. Specifically, for each new query, it retrieves the most similar past interactions and directly pastes them, \emph{verbatim}, into the prompt. DR can help the model see relevant input-output pairs but not directly codify any abstract or generalized solutions.\footnote{FH is similar to DR, but we include only a select (most relevant) input-output pairs in the memory content.}

Figure~\ref{fig:pseudocodes} (above) contains pseudocodes of all the primary methods and baselines considered in this paper.

\section{Experimental Setup}
\label{sec:experiments}

\subsection{Tasks and Datasets}

To rigorously evaluate DC's effectiveness, we focus on challenging tasks where contemporary state-of-the-art LLMs, such as GPT-4o and Claude 3.5, still face limitations. Rather than evaluating on benchmarks where performance is near saturation (e.g., BBH~\citep{suzgun2023challenging}, MGSM~\citep{shi2023language}, GSM8K~\citep{cobbe2021gsm8k}), we prioritize tasks that demand multi-step reasoning, heuristic search, strategic adaptation, and cumulative learning---that is, tasks in which iterative memory refinement can yield tangible improvements over time.\footnote{We release all the original input-output pairs in our codebase: \url{http://github.com/suzgunmirac/dynamic-cheatsheet}.}

Overall, the selected datasets include algorithmic, logical, and domain-specific reasoning tasks, each chosen to stress-test the model’s ability to refine its reasoning over time.

(a) \textbf{AIME 2020–2025 Exam Questions}: The American Invitational Mathematics Examination (AIME) is a prestigious high-school competition featuring complex problems across algebra, combinatorics, number theory, geometry, and probability. These questions require deep mathematical reasoning and multi-step problem-solving. We consider three subsets: AIME 2024\footnote{ \url{huggingface.co/datasets/HuggingFaceH4/aime_2024}} (30 questions), AIME 2025\footnote{\url{huggingface.co/datasets/yentinglin/aime_2025}.} (30 questions), and AIME 2020–2024\footnote{\mbox{\url{huggingface.co/datasets/di-zhang-fdu/AIME_1983_2024}}.} (133 questions).

(b) \textbf{GPQA-Diamond}~\citep{rein2024gpqa}: A high-quality, difficult subset of the Graduate-Level Google-Proof Q\&A (GPQA) benchmark, GPQA-Diamond contains 198 expert-validated questions across natural sciences, including biology, chemistry, and physics. These questions were correctly answered by domain experts but often missed by non-experts, making them ideal for evaluating DC’s ability to handle complex, multi-hop reasoning tasks.

(c) \textbf{Game of 24}~\citep{yao2023tree,suzgun2024metaprompting}: A heuristic-driven arithmetic challenge where the objective is to form an expression that evaluates to 24 using four given numbers exactly once. For instance, if the input values were ``7 7 8 11,'' one valid answer would be ``8*(7+7-11).'' This task  emphasizes systematic search, strategic reasoning, and pattern recognition. We use the 100 examples from \citep{suzgun2024metaprompting} to assess DC's capacity for refining computational heuristics and strategy over manual attempts.

(d) \textbf{Math Equation Balancer}: Focused on elementary arithmetic reasoning, this dataset requires the model to complete equations by inserting the appropriate operators  to form valid expressions. The task emphasizes the sequential placement of operators, as illustrated by the example ``1 ? 2 ? 3 = 6,'' where the model must identify the correct operators to satisfy the equation (``1 + 2 + 3 = 6'' or ``1 * 2 * 3 = 6''). We compiled 250 arithmetic expressions for this task. 

(e) \textbf{MMLU-Pro (Engineering and Physics)}~\citep{wang2024mmlupro}: A professional-level subset of the MMLU benchmark focused on physics and engineering. All questions are presented in a multiple-choice form. The original dataset contains 1,299 physics and 969 engineering questions. We sampled 250 questions from each subset.

\subsection{Language Models}

We evaluate the efficacy of DC across a range of language models. Our selection includes both state-of-the-art LLMs such as GPT-4o and Claude 3.5 Sonnet and their smaller-scale counterparts (namely, GPT-4o-mini and Claude 3.5 Haiku), as well as models such as DeepSeek R1 that are designed specifically for reasoning-intensive tasks.

\subsection{Evaluation Protocol}

To ensure standardized and reliable evaluation, all models are instructed to format their final answers in a structured, machine-readable format. All model answers are expected to be wrapped in the following XML-style tags:
\vspace{-1.0em}
\begin{minted}{html}
    <answer>
    (final answer)
    </answer>
\end{minted}
\vspace{-1.0em}

This explicit format ensures accurate and consistent parsing, eliminating errors arising from extraneous text or ambiguous outputs. Once extracted, the final answers are evaluated using their corresponding task-specific accuracy metric.

\subsubsection{Accuracy Metrics}
Given the diversity of the tasks, we use different accuracy metrics tailored to the specific requirements of each dataset.

\textbf{Soft Match (SM)} is a lenient metric that considers an answer correct if it matches the ground truth after ignoring minor formatting differences, such as punctuation or whitespace variations. We apply this metric to GPQA-Diamond, and MMLU Pro (Engineering and Physics), in which questions are presented in a multiple-choice format.

\textbf{Functionally Correct (FC)} is an even more flexible metric that evaluates whether the model’s output satisfies the task-specific constraints, even if the exact numeral presentation or formatting differs slightly from the reference solution. We apply this metric to the Game of 24, Math Equation Balancer, and AIME benchmarks.

\begin{table*}[!t]
\centering
\begin{tabular}{lccccc|cccccc}
\toprule
\multirow{2}{*}{\textbf{Tasks}} & \multicolumn{5}{c}{\textbf{Claude 3.5 Sonnet}} & \multicolumn{5}{c}{\textbf{GPT-4o}} \\
\cmidrule(lr){2-6} \cmidrule(lr){7-11}  
 & BL & DC-$\emptyset$ & DR & DC-Cu. & DC-RS & BL & DC-$\emptyset$ & DR & DC-Cu. & DC-RS  \\
\midrule
AIME 2024             & 23.3 & 36.7 & 43.3 & \textbf{50.0} & 46.7   & 20.0 & 36.7 & 26.7 & 36.7 & \textbf{40.0}  \\
AIME 2025             & 6.7  & 23.3 & 23.3 & \textbf{36.7} & 30.0   & 6.7  & 10.0 & 10.0 & 16.7 & \textbf{20.0}  \\
AIME 2020–24          & 6.7  & 30.1 & 39.1 & 38.4 & \textbf{40.6}  & 9.8  & 24.1 & 24.1 & 20.3  & \textbf{24.8}  \\
Game of 24            & 12.0 & 10.0 & 11.0 & \textbf{14.0} & \textbf{14.0}  & 10.0 & 19.0 & 6.0 & 93.0 & \textbf{99.0}  \\
GPQA Diamond          & 59.6 & 60.1 & 63.6 & 61.1 & \textbf{68.7}  & \textbf{57.1} & 57.1 & 55.1 & 58.1 & 57.1  \\
Math Eqn. Balancer  & 44.8 & 56.4 & 60.4 & \textbf{100} & 97.8  & 50.0 & 88.0 & \textbf{100} & \textbf{100} & 99.2   \\
MMLU Pro Eng.  & 61.2 & 57.2 & 65.2 & 66.8 & \textbf{67.6}  & \textbf{53.2} & 51.6 & 48.8 & 44.0  & 51.2  \\
MMLU Pro Physics      & 74.0 & 75.6 & 80.4 & 77.6 & \textbf{82.0}  & 75.6 & 70.8 & \textbf{75.6} & 70.4  & 75.2  \\
\bottomrule
\end{tabular}
\caption{
Performance comparison of Dynamic Cheatsheet (DC) variants for Claude 3.5 Sonnet and GPT-4o across multiple benchmarks. \textbf{BL} (Baseline): standard inference without memory;
\textbf{DC-$\emptyset$} (Empty Memory): includes structured problem-solving and explicit tool-use instructions but no memory retention mechanism; \textbf{DR} (Dynamic Retrieval): uses retrieval but lacks curated memory updates; \textbf{DC-Cu} (Cumulative Memory): iteratively accumulates model solutions but lacks retrieval; and \textbf{DC-RS} (Retrieval \& Synthesis): combines retrieval with memory refinement/synthesis. These results highlight substantial accuracy gains under DC: Claude 3.5 Sonnet's AIME 2024 accuracy jumps by 27\% under DC-Cu, and GPT-4o’s Game of 24 accuracy leaps from 10\% to 99\% under DC-RS.
}
\label{tab:dynamic_cheatsheet_results_merged}
\end{table*}

\section{Main Results}  
\label{sec:results}

\subsection{DC enables test-time learning and reduces repetitive errors}
\label{sec:game-of-24}

One of the most compelling illustrations of DC's capabilities emerges from the Game of 24 task. As seen in Table~\ref{tab:dynamic_cheatsheet_results_merged}, GPT-4o’s baseline accuracy on this arithmetic puzzle was just 10\%. Under DC-RS, its performance increased to 99\%, illustrating DC’s capacity for test-time learning and iterative refinement. Early in the task sequence, GPT-4o discovered a reliable, Python-based brute-force method to solve Game of 24 and later on recognized the repetitive structure of the problem. The model then encoded this approach into its  memory. Once established, GPT-4o consistently retrieved and applied the more or less same Python solution for subsequent examples, leading to rapid and accurate results.  

The performance under DC-$\emptyset$ (19\%) further highlights the positive impact of memory curation and retrieval. DC-$\emptyset$ uses the same core generator but keeps the memory empty, thus lacking the mechanism to store and reuse solutions. The large gap between 19\% (DC-$\emptyset$) and 99\% (DC-RS) confirms that effective memory usage, in which past solutions are retrieved and generalized, is the main driver of GPT-4o’s transformation from ad-hoc solver to near-perfect performer in Game of 24.

In contrast, Claude 3.5 Sonnet showed marginal gain, moving from 12\% to 14\%. Despite DC’s scaffolding, Claude did not internalize a generalized approach but instead continued to rely on manual arithmetic solutions. This underscores that while DC provides the framework for test-time adaptation, its ultimate success  hinges on the model’s innate capacity to identify and encode robust, reusable strategies.

\begin{figure}[ht]
\centering
\includegraphics[width=1\linewidth]{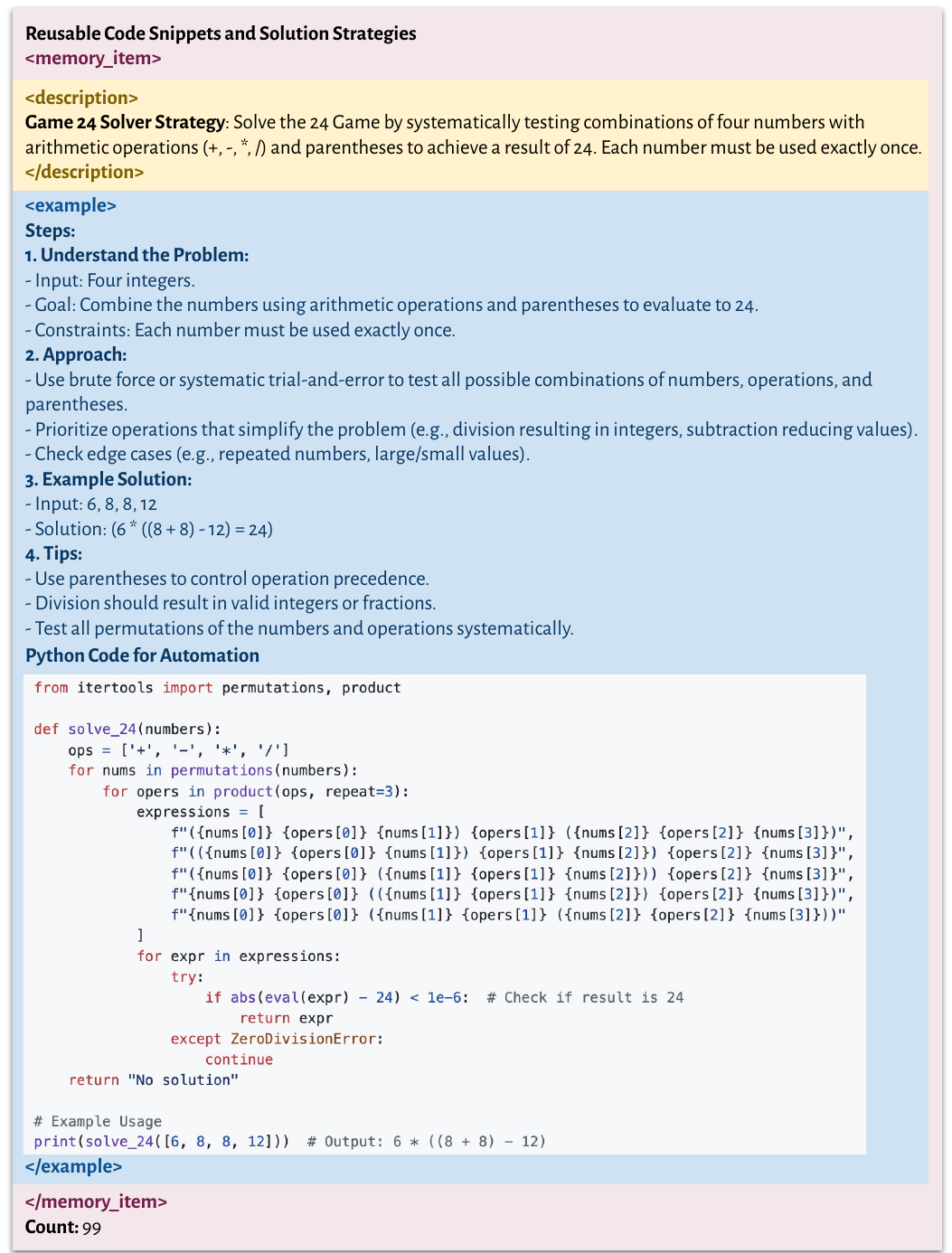}
\vspace{-0.3em}
\caption{
Excerpt from GPT-4o’s external memory after processing 100 examples from Game of 24 under DC-RS. Early in the test sequence, the model discovered a Python-based brute-force solution, stored it, and subsequently retrieved it for subsequent puzzles. This shift to structured code reuse resulted in a dramatic performance increase from 10\% to 99\% accuracy, eliminating arithmetic errors and redundant problem-solving efforts.}
\label{fig:gpt-4o-gameof24}
\end{figure}

\subsection{DC provides substantial improvements across various challenging reasoning benchmarks}
\label{sec:overall-results}

Beyond Game of 24, DC yielded significant gains across a range of complex mathematical and algorithmic tasks. \emph{See} Table~\ref{tab:dynamic_cheatsheet_results_merged}. The results below illustrate how iterative solution reuse can helpful in complex reasoning problems.

\textbf{AIME Exam Problems.}
The AIME exams provided some of the most dramatic improvements under DC. For Claude 3.5 Sonnet, performance on AIME 2020--2024 surged from 6.7\% to 40.6\% under DC-RS. A similar upward trend appeared on AIME 2024 (23.3\% to 50.0\%) and AIME 2025 (6.7\% to 36.7\%) under DC-Cu. DC-Cu, where the model curates memory after processing the input and does not involve a retrieval stage, also proved potent in recent exam sets, achieving highest accuracy scores in AIME 2024 and 2025. GPT-4o also showed some noteworthy gains. Its AIME 2024 performance raised from 20.0\% to 40.0\% under DC-RS, while its AIME 2025 score climbed from 6.7\% to 20.0\%. These boosts suggests that structured test-time-produced memory can help tackle difficult math problems. 

\textbf{GPQA-Diamond.} On GPQA-Diamond, Claude 3.5 Sonnet improved from 59.6\% to 68.7\% under DC-RS, a robust 9.1\% gain purely from test-time adaptation. DR (63.6\%) demonstrated that retrieval alone helps, but the further jump to 68.7\% highlights how memory curation and synthesis can yield additional benefits. By contrast, GPT-4o experienced only a slight increase from 57.1\% to 58.1\% with DC-RS; our quantitative analysis of the model's outputs and memory showed us that retrieval can, in some cases, introduce confusion, especially if suboptimal examples are recalled. This contrast between different models underscores how the success of retrieval-based adaptation partly depends on model-specific generation and curation capabilities.

\textbf{Math Equation Balancer.} As Table~\ref{tab:dynamic_cheatsheet_results_merged} shows, the baseline performance for Claude 3.5 Sonnet (44.8\%) rose to 98–100\% with DC-RS and DC-Cu, while GPT-4o similarly improved from 50.0\% to near-perfect accuracy (99–100\%). As observed in Game of 24, the models quickly learned an algorithmic or Python-based balancing routine, stored it in external memory, and repeatedly retrieved it, achieving exceptional consistency once the core method was established.

\textbf{MMLU-Pro Tasks.} For MMLU-Pro Eng. and Physics, Claude 3.5 Sonnet exhibited consistent gains, rising by up to 8.0\% in Physics (from 74\% to 82\%). Our examination of the curated memory entries shows that Claude temporarily stored and retrieved compact ``reference guides'' on engineering and physics principles, which might have proved beneficial for thematically similar questions. GPT-4o, on the other hand, observed slight decreases from the baseline on these tasks, suggesting that domain complexity and baseline knowledge gaps may attenuate DC’s benefits if curated memory is less reliable or consistent.

\subsection{Memory curation (DC) fosters generalization and provides gains over full-history-appending (FH)}

Whereas FH (full-history) simply appends every previous dialogue turn into the prompt, DC actively filters and synthesizes high-value content. As shown in Table~\ref{tab:full-history-results}, Sonnet under FH reached 26.7\% accuracy in 2024 questions, while DC-based methods hit 50.0\%. Similarly, GPT-4o managed a baseline of 20.0\% but fell to 6.7\% using FH, in direct contrast to 40.0\% with DC-RS. Excessive uncurated input-output pairs can not only overwhelm the model’s context window, dilute crucial insights and hamper retrieval efficiency, but also significantly increase inference costs over time. On the other hand, DC’s selective memory curation ensures that problem-solving tips or code snippets remain readily accessible without clutter, thus facilitating more robust and consistent improvements across consecutive queries.

\begin{table}[h]
\centering
\scalebox{0.88}{
\begin{tabular}{lccc|ccc}
\toprule
\multirow{2}{*}{\textbf{Tasks}} & \multicolumn{3}{c}{\textbf{Claude 3.5 Sonnet}} & \multicolumn{3}{c}{\textbf{GPT-4o}} \\
\cmidrule(lr){2-4} \cmidrule(lr){5-7}
& BL & FH & DC-Cu. & BL & FH & DC-RS \\
\midrule
AIME 2024 & 23.3 & 26.7 & \textbf{50.0} & 20.0 & 13.3 & \textbf{40.0} \\
AIME 2025 & 6.7 & 6.7 & \textbf{36.7} & 6.7 & 3.3 & \textbf{20.0} \\
\bottomrule
\end{tabular}
}
\caption{
Performance breakdown of \textbf{BL} (default baseline), \textbf{FH} (full history), \textbf{DC-Cu}, and \textbf{DC-RS} approaches under AIME 2024 and 2025. FH stores all past queries and outputs, while DC-Cu and DC-RS selectively refine stored memory. Results indicate that targeted memory curation in DC-RS leads to greater accuracy gains compared to full history retention, supporting the need for structured, self-updating knowledge mechanisms.}
\label{tab:full-history-results}
\end{table}

\subsection{DC fosters efficient tool usage / code generation}
\label{sec:fostering-tool-usage}

A successful behavior under DC is the LLMs’ inclination toward code generation to handle computationally intensive tasks. GPT-4o’s near-complete reliance on Python scripts for Game of 24 exemplifies this shift. Rather than performing manual arithmetic repeatedly, GPT-4o recognized that code-based brute force is more systematic. It generated, stored, and iteratively refined a Python function that tested permutations of numbers and operations, allowing it to solve each instance of Game of 24 with high accuracy.

This inclination toward automation illustrates DC’s potential to nurture efficient tool-usage: the capacity to recognize when external tools (e.g., Python, symbolic math engines, or dedicated solvers) are more robust than internally verbalized chain-of-thought calculations. While we restricted the scope of tool usage to Python interpreter in this study, future expansions could easily explore a broader suite of tools, potentially amplifying LLM performance in specialized domains such as computational biology or legal research.

\begin{figure}[!b]
\centering
\includegraphics[width=0.48\textwidth]{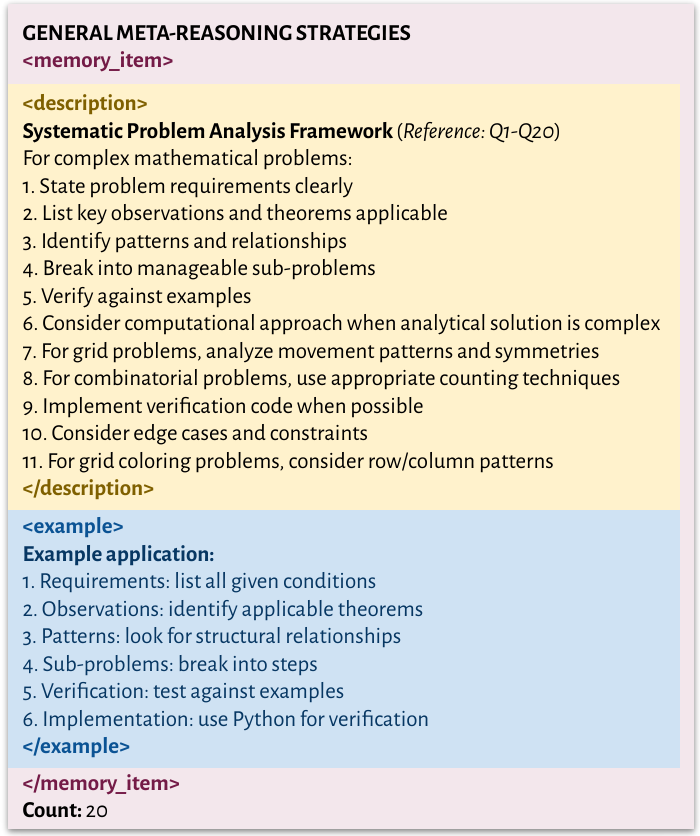}
\caption{
Example of Claude 3.5 Sonnet’s curated memory after processing 20 AIME 2024 questions under DC-Cu. The memory captures key solution strategies, enables the model to generalize across similar computational problems, and boosts its accuracy.
}
\label{fig:claude-sonnet-aime2024}
\end{figure}

\subsection{\mbox{Model scale and capacity impact DC effectiveness}}
\label{sec:model-scale-is-important}

\begin{figure*}[t]
\centering
\includegraphics[width=0.49\textwidth]{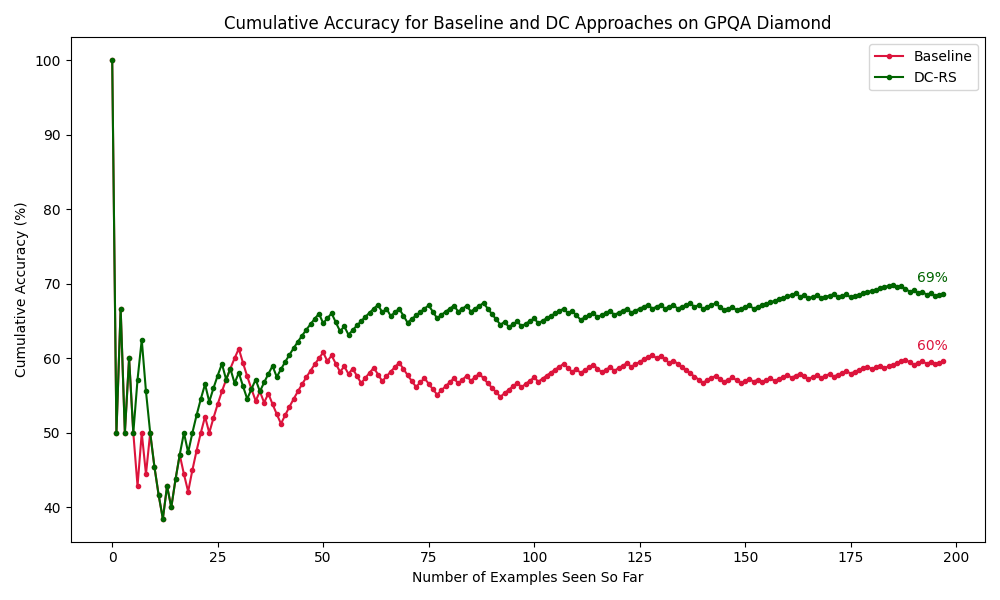}
\includegraphics[width=0.49\textwidth]{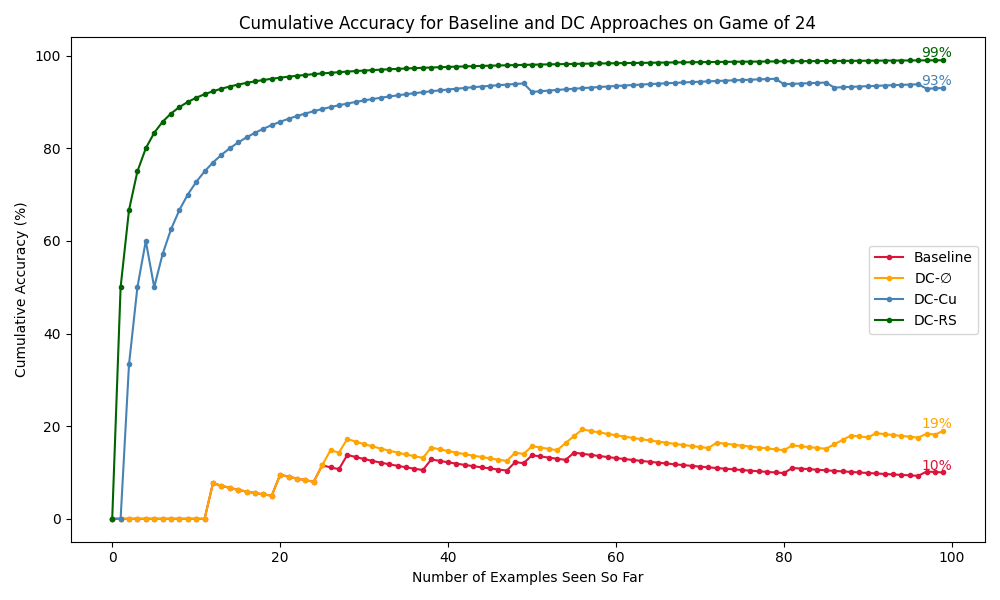}
\vspace{-0.3em}
\caption{
\emph{Cumulative} performance progression under DC for GPQA-Diamond (\emph{left}) and Game of 24 (\emph{right}). In GPQA-Diamond, Claude 3.5 Sonnet steadily improves as it accumulates relevant knowledge snippets (the first few points are noisy because $y$ measures cumulative accuracy). Meanwhile, in Game of 24, GPT-4o rapidly transitions from trial-and-error arithmetic to near-perfect performance once it recognizes and stores a Python-based solution. These trends highlight DC’s ability to enhance accuracy via iterative test-time learning. 
}
\label{fig:overview}
\end{figure*}

Our current results indicate that the effectiveness of DC is strongly tied to the model’s scale and underlying generative capacity. While Claude 3.5 Sonnet and GPT-4o showed notable gains across multiple tasks under DC, their smaller counterparts, Claude 3.5 Haiku and GPT-4o-mini, showed more limited and inconsistent gains. 

Table~\ref{tab:smaller-models-performance}, for instance, shows that Claude 3.5 Haiku achieved moderate gains under DC, with its accuracy on AIME 2024 rising from 10.0\% (baseline) to 36.7\% under DC-Cu. But gains on AIME 2025 were weaker, reaching only 13.3\% under DC-$\emptyset$ and DC-Cu. Interestingly, GPQA-Diamond saw an improvement from 43.4\% to 49.0\% under DC-RS, suggesting that retrieval-based adaptation might still provide utility in smaller models.

\begin{table}[h]
\centering
\scalebox{0.92}{
\begin{tabular}{lcccc}
\toprule
\multirow{2}{*}{\textbf{Tasks}} & \multicolumn{4}{c}{\textbf{Claude 3.5 Haiku}} \\
\cmidrule(lr){2-5}
& BL & DC-$\emptyset$ & DC-Cu. & DC-RS \\
\midrule
AIME 2024 & 10.0 & 26.7 & \textbf{36.7} & 30.0 \\
AIME 2025 & 0.0 & 13.3 & 13.3 & 10.0 \\
GPQA-Diamond & 43.4 & 41.9 & 43.7 & \textbf{49.0}\\
\bottomrule
\end{tabular}
}
\scalebox{0.92}{
\begin{tabular}{lcccc}
\toprule
\multirow{2}{*}{\textbf{Tasks}} & \multicolumn{4}{c}{\textbf{GPT-4o-mini}} \\
\cmidrule(lr){2-5} 
& BL & DC-$\emptyset$ & DC-Cu. & DC-RS \\
\midrule
AIME 2024 & 16.7 & \textbf{20.0} & 13.3 & 13.3 \\
AIME 2025 & 10.0 & \textbf{13.3} & \textbf{13.3} & \textbf{16.7} \\
GPQA-Diamond & \textbf{34.3} & \textbf{34.3} & 33.8 & 32.3 \\
\bottomrule
\end{tabular}
}
\vspace{-0.1em}
\caption{Performance of Claude 3.5 Haiku and GPT-4o-mini, the smaller counterparts of Claude 3.5 Sonnet and GPT-4o, across AIME (2024, 2025) and GPQA-Diamond. These smaller models struggle to fully leverage DC, suggesting that memory-based adaptation is most effective when the base LM has sufficient generative competence. Performance improvements are more muted, highlighting the dependency of DC on model-scale reasoning ability.}
\label{tab:smaller-models-performance}
\vspace{0.3em}
\end{table}

That said, GPT-4o-mini (Table~\ref{tab:smaller-models-performance}) showed even smaller gains, with some variants leading to slight declines in performance. On AIME 2024, DC-$\emptyset$ provided a 20.0\% boost, but both DC-Cu and DC-RS performed worse than baseline. AIME 2025 showed a minor improvement, peaking at 16.7\% under DC-RS. On GPQA-Diamond, GPT-4o-mini’s performance, however, remained largely stagnant or slightly declined under memory-based adaptation, suggesting that it struggled to leverage stored information effectively.

These imply two drawbacks of smaller models under DC:

\textbf{(a) \emph{Generative competence}.} For DC to be effective, the base model must produce correct solutions with sufficient frequency to populate the memory with high-quality, reusable strategies. Smaller models, such as GPT-4o-mini and Claude 3.5 Haiku, generate correct solutions less reliably, leading to a sparse or low-quality memory repository. As a result, iterative refinement stalls because the stored knowledge consists mostly of incorrect or partial attempts.

\textbf{(b) \emph{Contextual and memory curation limitations}.} Smaller models struggle with long-context understanding/generation and memory retrieval, leading to inefficient or irrelevant memory usage. Unlike their larger counterparts, which can more effectively retrieve and synthesize solutions from stored heuristics, smaller models often fail to retrieve the most relevant past solutions or misapply retrieved knowledge to new problems. This results in inconsistent performance under DC-RS, particularly in tasks requiring complex reasoning or strategic adaptation.

\subsection{Test-time task similarity and example ordering can amplify DC’s overall impact}
\label{sec:task-similarity-and-ordering}

Another central insight is that DC thrives when test examples share structural similarities. In both Game of 24 and Math Equation Balancer, once GPT-4o identified an efficient solution, it reused it consistently for subsequent tasks. Similarly, in AIME, discovering a geometry or combinatorics strategy allowed for easy transfer across questions of analogous structure. Consequently, tasks arranged to present \emph{related} questions early may accelerate and improve the model’s test-time learning. This suggests that \emph{curriculum-style} learning~\citep{bengio2009curriculum}, where simpler or archetypal problems are presented first to build a repository of valid heuristics, may potentially bootstrap performance. \emph{Cf.}~\citep{lopez2017gradient,zelikman2022star,chen2024self}

\section{Additional Analyses and Discussions}
\label{sec:discussion}

\textbf{Reasoning and information efficiency.}
One key insight is that DC reduces the need to ``reinvent the wheel'' for each query. By encoding and reusing well-established techniques (e.g., Python-based solving for Game of 24), models can bypass repeated rediscovery of the same strategies. This significantly cuts down reasoning overhead and token usage in subsequent queries, though the initial cost of discovering a robust approach and curating it remains non-trivial.

\textbf{DC performs better than majority voting (MV).} To test if DC provides advantages over conventional MV at inference, we also tested Sonnet on AIME 2024 and 2025 using both approaches. MV, which selects the most common answer from three independent generations, yielded no improvements over single-shot inference. As seen in Table~\ref{tab:majority-voting-results}, on AIME 2024, MV performed identically to the baseline (23.3\%), while on AIME 2025, it remained at 6.7\%, offering no tangible gain. Even with DC-$\emptyset$, MV slightly underperformed (33.3\% vs. 36.7\%). In contrast, DC-Cu outperformed MV, reaching 50.0\% on AIME 2024 and 36.7\% on AIME 2025. Unlike MV, which \emph{passively} aggregates outputs, DC \emph{actively} refines knowledge over time, eliminating errors and improving solution quality. This confirms that memory-based adaptation is far more effective than simple statistical voting in complex reasoning tasks.

\begin{table}[h]
\centering
\scalebox{0.86}{
\begin{tabular}{lcc|cc|c}
\toprule
\multirow{2}{*}{\textbf{Tasks}} & \multicolumn{5}{c}{\textbf{Claude 3.5 Sonnet}} \\
\cmidrule(lr){2-6}
& BL & MV(BL) & DC-$\emptyset$ & MV(DC-$\emptyset$) & DC-Cu.  \\
\midrule
AIME 2024 & 23.3 & 23.33 & 36.7 & 33.3 & \textbf{50.0}\\
AIME 2025 & 6.7 & 6.7 & 23.3 & 23.3 & \textbf{36.7} \\
\bottomrule
\end{tabular}
}
\caption{Comparison of majority voting (MV) with DC on AIME.
\vspace{0.1em}
}
\label{tab:majority-voting-results}
\end{table}

\textbf{Clustering of errors and corrections.}
Our experiments suggest that errors and their corrections often cluster in a latent embedding space. \emph{See}~Figure~\ref{fig:question-embeddings}. Once a model acquires a high-quality heuristic for a cluster of related queries, it can apply this knowledge to tightly embedded neighbors. However, faulty heuristics that slip into memory can be equally amplified. Ensuring that the memory remains ``clean'' thus requires careful curation and, if necessary, pruning to avoid propagating erroneous strategies.

\textbf{Transferability of memory content across models.}
We also observed that larger models, such as Claude 3.5 Sonnet and GPT-4o, can sometimes produce higher-quality strategies that, in principle, could benefit smaller models if the memory is transferred. However, if a smaller model lacks the generative capacity to interpret or refine those strategies correctly, its performance can stall or degrade. In our ablation experiments, we observed mixed results. This indicates that memory entries, while helpful, cannot fully compensate for inadequate base capability.

\textbf{Long-context generation versus understanding.}
Most large LLMs excel at \emph{processing} lengthy inputs but struggle to \emph{generate} comparably long\footnote{\emph{See, e.g.,} \citep{liu2024longgenbench}.} and well-organized outputs. DC’s memory curation after each query can demand precise reproduction or modification of prior knowledge. We observed instances where the model merely references or abbreviates the existing memory (e.g., “Previous content [...] preserved”) instead of explicitly rewriting it. Such truncated memory updates can reduce the quality of stored heuristics over time. Potential solutions include maintaining a structured, external database that the LM can reference without regenerating large swaths of text each time.

\textbf{Retrieval bottlenecks and noise.}
While retrieval-based variants (e.g., DC-RS) can substantially improve accuracy, poorly filtered retrieval mechanisms can introduce confusion, particularly when presented with highly diverse or loosely related queries. For example, in our experiments, GPT-4o’s performance occasionally dipped in GPQA-Diamond due to suboptimal retrieval choices. This underscores the importance of robust retrieval methods (e.g., dense vector search, advanced ranking algorithms) that can reliably surface higher quality exemplars or heuristics while suppressing irrelevant or contradictory texts.

\textbf{Hierarchical and modular memory.}
As LLM deployments scale, specialized domains may benefit from subdividing or hierarchically organizing memory. For instance, a system could maintain separate curated memories for topics like combinatorics or physics, each updated by a specialized retrieval or curation mechanism. This may reduce the load on a unified memory store and help isolate errors within their respective domains, with the goal of further improving the clarity and reliability of retrieved heuristics.

\textbf{Time and token complexity.}
Although DC requires memory curation after each query, it optimizes efficiency over time by reducing redundant computation and token usage.\footnote{On AIME 2024, Claude Sonnet averaged 370 tokens under BL, 494 under DC-$\emptyset$, 1035 under DC-RS, and 1831 under DC-Cu.} As the model retrieves and refines solutions, memory maintenance becomes a net gain rather than a cost. However, its sequential structure still poses challenges for large-scale parallel or batch tasks requiring independent inference.

\textbf{Smaller or more specialized models and R1 experiments.}
Finally, we note that smaller models, such as GPT-4o-mini, show limited gains under DC, as seen in Table~\ref{tab:smaller-models-performance}. Additional experiments with ``R1'' models such as DeepSeek R1 and o1 similarly showed minimal or inconsistent improvements. In these cases, these models' generative ability appears too restricted to produce reliable strategies for storage or to interpret retrieved heuristics effectively. The solutions were far too verbose and long. Without sufficiently accurate and efficient base solutions, memory curation cannot yield substantial gains. This limitation ties back to the core premise that effective DC demands a capable foundation model to seed and refine the curated knowledge.

Overall, DC offers a useful and practical framework for continuous, test-time learning in LLMs. Our findings emphasize the synergy between model capacity and memory curation, the importance of structural task similarity and retrieval precision, and the benefits of offloading repeated computations to flexible external stores (e.g., Python scripts). At the same time, alternative mechanisms (e.g., specialized sub-memories or adaptive example ordering) and more sophisticated retrieval techniques (e.g., topological clustering) remain promising directions for further research.

\section*{Acknowledgments}
We thank
Batu El,
Sabri Eyuboglu,
Tayfun Gur,
Emily Shen,
Jake Silberg,
Elana Simon, and
Kyle Swanson for
their helpful comments and suggestions. We also thank the members of the James Zou Lab at Stanford for their feedback in the early stages of this project.
Suzgun gratefully acknowledges the support of an HAI-SAP Fellowship.

\bibliography{example_paper}

\begin{thebibliography}{78}
\providecommand{\natexlab}[1]{#1}
\providecommand{\url}[1]{\texttt{#1}}
\expandafter\ifx\csname urlstyle\endcsname\relax
  \providecommand{\doi}[1]{doi: #1}\else
  \providecommand{\doi}{doi: \begingroup \urlstyle{rm}\Url}\fi

\bibitem[Amari(1998)]{amari1998natural}
Amari, S.-I.
\newblock Natural gradient works efficiently in learning.
\newblock \emph{Neural computation}, 10\penalty0 (2):\penalty0 251--276, 1998.

\bibitem[Arcuschin et~al.(2025)Arcuschin, Janiak, Krzyzanowski, Rajamanoharan, Nanda, and Conmy]{arcuschin2025chainofthought}
Arcuschin, I., Janiak, J., Krzyzanowski, R., Rajamanoharan, S., Nanda, N., and Conmy, A.
\newblock Chain-of-thought reasoning in the wild is not always faithful.
\newblock In \emph{Workshop on Reasoning and Planning for Large Language Models}, 2025.
\newblock URL \url{https://openreview.net/forum?id=L8094Whth0}.

\bibitem[Asai et~al.(2023)Asai, Wu, Wang, Sil, and Hajishirzi]{asai2023self}
Asai, A., Wu, Z., Wang, Y., Sil, A., and Hajishirzi, H.
\newblock Self-rag: Learning to retrieve, generate, and critique through self-reflection.
\newblock In \emph{The Twelfth International Conference on Learning Representations}, 2023.

\bibitem[Bengio et~al.(2009)Bengio, Louradour, Collobert, and Weston]{bengio2009curriculum}
Bengio, Y., Louradour, J., Collobert, R., and Weston, J.
\newblock Curriculum learning.
\newblock In \emph{Proceedings of the 26th annual international conference on machine learning}, pp.\  41--48, 2009.

\bibitem[Besta et~al.(2024)Besta, Blach, Kubicek, Gerstenberger, Podstawski, Gianinazzi, Gajda, Lehmann, Niewiadomski, Nyczyk, et~al.]{besta2024graph}
Besta, M., Blach, N., Kubicek, A., Gerstenberger, R., Podstawski, M., Gianinazzi, L., Gajda, J., Lehmann, T., Niewiadomski, H., Nyczyk, P., et~al.
\newblock Graph of thoughts: Solving elaborate problems with large language models.
\newblock In \emph{Proceedings of the AAAI Conference on Artificial Intelligence}, volume~38, pp.\  17682--17690, 2024.

\bibitem[Borgeaud et~al.(2022)Borgeaud, Mensch, Hoffmann, Cai, Rutherford, Millican, Van Den~Driessche, Lespiau, Damoc, Clark, et~al.]{borgeaud2022improving}
Borgeaud, S., Mensch, A., Hoffmann, J., Cai, T., Rutherford, E., Millican, K., Van Den~Driessche, G.~B., Lespiau, J.-B., Damoc, B., Clark, A., et~al.
\newblock Improving language models by retrieving from trillions of tokens.
\newblock In \emph{International conference on machine learning}, pp.\  2206--2240. PMLR, 2022.

\bibitem[Bottou \& Cun(2003)Bottou and Cun]{bottou2003large}
Bottou, L. and Cun, Y.
\newblock Large scale online learning.
\newblock \emph{Advances in neural information processing systems}, 16, 2003.

\bibitem[Bottou \& Le~Cun(2005)Bottou and Le~Cun]{bottou2005online}
Bottou, L. and Le~Cun, Y.
\newblock On-line learning for very large data sets.
\newblock \emph{Applied stochastic models in business and industry}, 21\penalty0 (2):\penalty0 137--151, 2005.

\bibitem[Boudiaf et~al.(2022)Boudiaf, Mueller, Ben~Ayed, and Bertinetto]{boudiaf2022parameter}
Boudiaf, M., Mueller, R., Ben~Ayed, I., and Bertinetto, L.
\newblock Parameter-free online test-time adaptation.
\newblock In \emph{Proceedings of the IEEE/CVF Conference on Computer Vision and Pattern Recognition}, pp.\  8344--8353, 2022.

\bibitem[Bulatov et~al.(2022)Bulatov, Kuratov, and Burtsev]{bulatov2022recurrent}
Bulatov, A., Kuratov, Y., and Burtsev, M.
\newblock Recurrent memory transformer.
\newblock \emph{Advances in Neural Information Processing Systems}, 35:\penalty0 11079--11091, 2022.

\bibitem[Chen et~al.(2024)Chen, Deng, Yuan, Ji, and Gu]{chen2024self}
Chen, Z., Deng, Y., Yuan, H., Ji, K., and Gu, Q.
\newblock Self-play fine-tuning converts weak language models to strong language models.
\newblock \emph{arXiv preprint arXiv:2401.01335}, 2024.

\bibitem[Cobbe et~al.(2021)Cobbe, Kosaraju, Bavarian, Chen, Jun, Kaiser, Plappert, Tworek, Hilton, Nakano, Hesse, and Schulman]{cobbe2021gsm8k}
Cobbe, K., Kosaraju, V., Bavarian, M., Chen, M., Jun, H., Kaiser, L., Plappert, M., Tworek, J., Hilton, J., Nakano, R., Hesse, C., and Schulman, J.
\newblock Training verifiers to solve math word problems.
\newblock \emph{arXiv preprint arXiv:2110.14168}, 2021.

\bibitem[Feng et~al.(2024)Feng, Han, Lin, Liu, and You]{feng2024thoughtretriever}
Feng, T., Han, P., Lin, G., Liu, G., and You, J.
\newblock Thought-retriever: Don{\textquoteright}t just retrieve raw data, retrieve thoughts, 2024.
\newblock URL \url{https://openreview.net/forum?id=SkDNQbMQba}.

\bibitem[Feng et~al.(2022)Feng, Li, Song, Zheng, and Koehn]{feng2022learn}
Feng, Y., Li, F., Song, Z., Zheng, B., and Koehn, P.
\newblock Learn to remember: Transformer with recurrent memory for document-level machine translation.
\newblock \emph{arXiv preprint arXiv:2205.01546}, 2022.

\bibitem[Golovneva et~al.(2023)Golovneva, O'Brien, Pasunuru, Wang, Zettlemoyer, Fazel-Zarandi, and Celikyilmaz]{golovneva2023pathfinder}
Golovneva, O., O'Brien, S., Pasunuru, R., Wang, T., Zettlemoyer, L., Fazel-Zarandi, M., and Celikyilmaz, A.
\newblock Pathfinder: Guided search over multi-step reasoning paths.
\newblock \emph{arXiv preprint arXiv:2312.05180}, 2023.

\bibitem[Gou et~al.(2023)Gou, Shao, Gong, Shen, Yang, Duan, and Chen]{gou2023critic}
Gou, Z., Shao, Z., Gong, Y., Shen, Y., Yang, Y., Duan, N., and Chen, W.
\newblock Critic: Large language models can self-correct with tool-interactive critiquing.
\newblock \emph{arXiv preprint arXiv:2305.11738}, 2023.

\bibitem[Graves(2013)]{graves2013generating}
Graves, A.
\newblock Generating sequences with recurrent neural networks.
\newblock \emph{arXiv preprint arXiv:1308.0850}, 2013.

\bibitem[Graves et~al.(2014)Graves, Wayne, and Danihelka]{graves2014neural}
Graves, A., Wayne, G., and Danihelka, I.
\newblock Neural turing machines.
\newblock \emph{arXiv preprint arXiv:1410.5401}, 2014.

\bibitem[Gururangan et~al.(2020)Gururangan, Marasovi{\'c}, Swayamdipta, Lo, Beltagy, Downey, and Smith]{gururangan2020don}
Gururangan, S., Marasovi{\'c}, A., Swayamdipta, S., Lo, K., Beltagy, I., Downey, D., and Smith, N.~A.
\newblock Don't stop pretraining: Adapt language models to domains and tasks.
\newblock \emph{arXiv preprint arXiv:2004.10964}, 2020.

\bibitem[Guu et~al.(2020)Guu, Lee, Tung, Pasupat, and Chang]{guu2020retrieval}
Guu, K., Lee, K., Tung, Z., Pasupat, P., and Chang, M.
\newblock Retrieval augmented language model pre-training.
\newblock In \emph{International conference on machine learning}, pp.\  3929--3938. PMLR, 2020.

\bibitem[He et~al.(2024)He, Karlinsky, Kim, McAuley, Krotov, and Feris]{he2024camelot}
He, Z., Karlinsky, L., Kim, D., McAuley, J., Krotov, D., and Feris, R.
\newblock Camelot: Towards large language models with training-free consolidated associative memory.
\newblock \emph{arXiv preprint arXiv:2402.13449}, 2024.

\bibitem[Joulin \& Mikolov(2015)Joulin and Mikolov]{joulin2015inferring}
Joulin, A. and Mikolov, T.
\newblock Inferring algorithmic patterns with stack-augmented recurrent nets.
\newblock \emph{Advances in neural information processing systems}, 28, 2015.

\bibitem[Karpicke \& Blunt(2011)Karpicke and Blunt]{karpicke2011retrieval}
Karpicke, J.~D. and Blunt, J.~R.
\newblock Retrieval practice produces more learning than elaborative studying with concept mapping.
\newblock \emph{Science}, 331\penalty0 (6018):\penalty0 772--775, 2011.

\bibitem[Karpicke \& Roediger~III(2008)Karpicke and Roediger~III]{karpicke2008critical}
Karpicke, J.~D. and Roediger~III, H.~L.
\newblock The critical importance of retrieval for learning.
\newblock \emph{science}, 319\penalty0 (5865):\penalty0 966--968, 2008.

\bibitem[Karpukhin et~al.(2020)Karpukhin, Oguz, Min, Lewis, Wu, Edunov, Chen, and Yih]{karpukhin2020dense}
Karpukhin, V., Oguz, B., Min, S., Lewis, P.~S., Wu, L., Edunov, S., Chen, D., and Yih, W.-t.
\newblock Dense passage retrieval for open-domain question answering.
\newblock In \emph{EMNLP (1)}, pp.\  6769--6781, 2020.

\bibitem[Khandelwal et~al.(2020)Khandelwal, Levy, Jurafsky, Zettlemoyer, and Lewis]{Khandelwal2020Generalization}
Khandelwal, U., Levy, O., Jurafsky, D., Zettlemoyer, L., and Lewis, M.
\newblock Generalization through memorization: Nearest neighbor language models.
\newblock In \emph{International Conference on Learning Representations}, 2020.
\newblock URL \url{https://openreview.net/forum?id=HklBjCEKvH}.

\bibitem[Kojima et~al.(2022)Kojima, Gu, Reid, Matsuo, and Iwasawa]{kojima2022large}
Kojima, T., Gu, S.~S., Reid, M., Matsuo, Y., and Iwasawa, Y.
\newblock Large language models are zero-shot reasoners.
\newblock \emph{Advances in neural information processing systems}, 35:\penalty0 22199--22213, 2022.

\bibitem[Krause et~al.(2019)Krause, Kahembwe, Murray, and Renals]{krause2019dynamic}
Krause, B., Kahembwe, E., Murray, I., and Renals, S.
\newblock Dynamic evaluation of transformer language models.
\newblock \emph{arXiv preprint arXiv:1904.08378}, 2019.

\bibitem[Lazaridou et~al.(2023)Lazaridou, Gribovskaya, Stokowiec, and Grigorev]{lazaridou2023internetaugmented}
Lazaridou, A., Gribovskaya, E., Stokowiec, W.~J., and Grigorev, N.
\newblock Internet-augmented language models through few-shot prompting for open-domain question answering, 2023.
\newblock URL \url{https://openreview.net/forum?id=hFCUPkSSRE}.

\bibitem[Lewis et~al.(2020)Lewis, Perez, Piktus, Petroni, Karpukhin, Goyal, K{\"u}ttler, Lewis, Yih, Rockt{\"a}schel, et~al.]{lewis2020retrieval}
Lewis, P., Perez, E., Piktus, A., Petroni, F., Karpukhin, V., Goyal, N., K{\"u}ttler, H., Lewis, M., Yih, W.-t., Rockt{\"a}schel, T., et~al.
\newblock Retrieval-augmented generation for knowledge-intensive nlp tasks.
\newblock \emph{Advances in neural information processing systems}, 33:\penalty0 9459--9474, 2020.

\bibitem[Liu et~al.(2024{\natexlab{a}})Liu, Lin, Hewitt, Paranjape, Bevilacqua, Petroni, and Liang]{liu2024lost}
Liu, N.~F., Lin, K., Hewitt, J., Paranjape, A., Bevilacqua, M., Petroni, F., and Liang, P.
\newblock Lost in the middle: How language models use long contexts.
\newblock \emph{Transactions of the Association for Computational Linguistics}, 12:\penalty0 157--173, 2024{\natexlab{a}}.

\bibitem[Liu et~al.(2024{\natexlab{b}})Liu, Dong, Hu, and Chu]{liu2024longgenbench}
Liu, X., Dong, P., Hu, X., and Chu, X.
\newblock Longgenbench: Long-context generation benchmark.
\newblock \emph{arXiv preprint arXiv:2410.04199}, 2024{\natexlab{b}}.

\bibitem[Liu et~al.(2021)Liu, Kothari, Van~Delft, Bellot-Gurlet, Mordan, and Alahi]{liu2021tttplus}
Liu, Y., Kothari, P., Van~Delft, B., Bellot-Gurlet, B., Mordan, T., and Alahi, A.
\newblock Ttt++: When does self-supervised test-time training fail or thrive?
\newblock \emph{Advances in Neural Information Processing Systems}, 34:\penalty0 21808--21820, 2021.

\bibitem[Long(2023)]{long2023large}
Long, J.
\newblock Large language model guided tree-of-thought.
\newblock \emph{arXiv preprint arXiv:2305.08291}, 2023.

\bibitem[Lopez-Paz \& Ranzato(2017)Lopez-Paz and Ranzato]{lopez2017gradient}
Lopez-Paz, D. and Ranzato, M.
\newblock Gradient episodic memory for continual learning.
\newblock \emph{Advances in neural information processing systems}, 30, 2017.

\bibitem[Lu et~al.(2023)Lu, Peng, Cheng, Galley, Chang, Wu, Zhu, and Gao]{lu2023chameleon}
Lu, P., Peng, B., Cheng, H., Galley, M., Chang, K.-W., Wu, Y.~N., Zhu, S.-C., and Gao, J.
\newblock Chameleon: Plug-and-play compositional reasoning with large language models.
\newblock \emph{Advances in Neural Information Processing Systems}, 36:\penalty0 43447--43478, 2023.

\bibitem[Madaan et~al.(2022)Madaan, Tandon, Clark, and Yang]{madaan2022memory}
Madaan, A., Tandon, N., Clark, P., and Yang, Y.
\newblock Memory-assisted prompt editing to improve gpt-3 after deployment.
\newblock In \emph{Proceedings of the 2022 Conference on Empirical Methods in Natural Language Processing}, pp.\  2833--2861, 2022.

\bibitem[Madaan et~al.(2023)Madaan, Tandon, Gupta, Hallinan, Gao, Wiegreffe, Alon, Dziri, Prabhumoye, Yang, et~al.]{madaan2023self}
Madaan, A., Tandon, N., Gupta, P., Hallinan, S., Gao, L., Wiegreffe, S., Alon, U., Dziri, N., Prabhumoye, S., Yang, Y., et~al.
\newblock Self-refine: Iterative refinement with self-feedback.
\newblock \emph{Advances in Neural Information Processing Systems}, 36:\penalty0 46534--46594, 2023.

\bibitem[McCloskey \& Cohen(1989)McCloskey and Cohen]{mccloskey1989catastrophic}
McCloskey, M. and Cohen, N.~J.
\newblock Catastrophic interference in connectionist networks: The sequential learning problem.
\newblock In \emph{Psychology of learning and motivation}, volume~24, pp.\  109--165. Elsevier, 1989.

\bibitem[Mikolov et~al.(2010)Mikolov, Karafi{\'a}t, Burget, Cernock{\`y}, and Khudanpur]{mikolov2010recurrent}
Mikolov, T., Karafi{\'a}t, M., Burget, L., Cernock{\`y}, J., and Khudanpur, S.
\newblock Recurrent neural network based language model.
\newblock In \emph{Interspeech}, volume~2, pp.\  1045--1048. Makuhari, 2010.

\bibitem[Munkhdalai et~al.(2019)Munkhdalai, Sordoni, Wang, and Trischler]{munkhdalai2019metalearned}
Munkhdalai, T., Sordoni, A., Wang, T., and Trischler, A.
\newblock Metalearned neural memory.
\newblock \emph{Advances in Neural Information Processing Systems}, 32, 2019.

\bibitem[Niu et~al.(2022)Niu, Wu, Zhang, Chen, Zheng, Zhao, and Tan]{niu2022efficient}
Niu, S., Wu, J., Zhang, Y., Chen, Y., Zheng, S., Zhao, P., and Tan, M.
\newblock Efficient test-time model adaptation without forgetting.
\newblock In \emph{International conference on machine learning}, pp.\  16888--16905. PMLR, 2022.

\bibitem[Qin et~al.(2023)Qin, Liang, Ye, Zhu, Yan, Lu, Lin, Cong, Tang, Qian, et~al.]{qin2023toolllm}
Qin, Y., Liang, S., Ye, Y., Zhu, K., Yan, L., Lu, Y., Lin, Y., Cong, X., Tang, X., Qian, B., et~al.
\newblock Toolllm: Facilitating large language models to master 16000+ real-world apis.
\newblock \emph{arXiv preprint arXiv:2307.16789}, 2023.

\bibitem[Rannen-Triki et~al.(2024)Rannen-Triki, Bornschein, Pascanu, Hutter, Gy{\"o}rgy, Galashov, Teh, and Titsias]{rannen2024revisiting}
Rannen-Triki, A., Bornschein, J., Pascanu, R., Hutter, M., Gy{\"o}rgy, A., Galashov, A., Teh, Y.~W., and Titsias, M.~K.
\newblock Revisiting dynamic evaluation: Online adaptation for large language models.
\newblock \emph{arXiv preprint arXiv:2403.01518}, 2024.

\bibitem[Rein et~al.(2024)Rein, Hou, Stickland, Petty, Pang, Dirani, Michael, and Bowman]{rein2024gpqa}
Rein, D., Hou, B.~L., Stickland, A.~C., Petty, J., Pang, R.~Y., Dirani, J., Michael, J., and Bowman, S.~R.
\newblock {GPQA}: A graduate-level google-proof q\&a benchmark.
\newblock In \emph{First Conference on Language Modeling}, 2024.
\newblock URL \url{https://openreview.net/forum?id=Ti67584b98}.

\bibitem[Roediger \& Butler(2011)Roediger and Butler]{roediger2011critical}
Roediger, H.~L. and Butler, A.~C.
\newblock The critical role of retrieval practice in long-term retention.
\newblock \emph{Trends in cognitive sciences}, 15\penalty0 (1):\penalty0 20--27, 2011.

\bibitem[Schick et~al.(2023)Schick, Dwivedi-Yu, Dess{\`\i}, Raileanu, Lomeli, Hambro, Zettlemoyer, Cancedda, and Scialom]{schick2023toolformer}
Schick, T., Dwivedi-Yu, J., Dess{\`\i}, R., Raileanu, R., Lomeli, M., Hambro, E., Zettlemoyer, L., Cancedda, N., and Scialom, T.
\newblock Toolformer: Language models can teach themselves to use tools.
\newblock \emph{Advances in Neural Information Processing Systems}, 36:\penalty0 68539--68551, 2023.

\bibitem[Shen et~al.(2023)Shen, Song, Tan, Li, Lu, and Zhuang]{shen2023hugginggpt}
Shen, Y., Song, K., Tan, X., Li, D., Lu, W., and Zhuang, Y.
\newblock Hugging{GPT}: Solving {AI} tasks with chat{GPT} and its friends in hugging face.
\newblock In \emph{Thirty-seventh Conference on Neural Information Processing Systems}, 2023.
\newblock URL \url{https://openreview.net/forum?id=yHdTscY6Ci}.

\bibitem[Shi et~al.(2022)Shi, Fried, Ghazvininejad, Zettlemoyer, and Wang]{shi2022natural}
Shi, F., Fried, D., Ghazvininejad, M., Zettlemoyer, L., and Wang, S.~I.
\newblock Natural language to code translation with execution.
\newblock In \emph{Proceedings of the 2022 Conference on Empirical Methods in Natural Language Processing}, pp.\  3533--3546, 2022.

\bibitem[Shi et~al.(2023)Shi, Suzgun, Freitag, Wang, Srivats, Vosoughi, Chung, Tay, Ruder, Zhou, Das, and Wei]{shi2023language}
Shi, F., Suzgun, M., Freitag, M., Wang, X., Srivats, S., Vosoughi, S., Chung, H.~W., Tay, Y., Ruder, S., Zhou, D., Das, D., and Wei, J.
\newblock Language models are multilingual chain-of-thought reasoners.
\newblock In \emph{The Eleventh International Conference on Learning Representations}, 2023.
\newblock URL \url{https://openreview.net/forum?id=fR3wGCk-IXp}.

\bibitem[Shi et~al.(2024)Shi, Min, Yasunaga, Seo, James, Lewis, Zettlemoyer, and Yih]{shietal2024replug}
Shi, W., Min, S., Yasunaga, M., Seo, M., James, R., Lewis, M., Zettlemoyer, L., and Yih, W.-t.
\newblock {REPLUG}: Retrieval-augmented black-box language models.
\newblock In Duh, K., Gomez, H., and Bethard, S. (eds.), \emph{Proceedings of the 2024 Conference of the North American Chapter of the Association for Computational Linguistics: Human Language Technologies (Volume 1: Long Papers)}, pp.\  8371--8384, Mexico City, Mexico, June 2024. Association for Computational Linguistics.
\newblock \doi{10.18653/v1/2024.naacl-long.463}.
\newblock URL \url{https://aclanthology.org/2024.naacl-long.463/}.

\bibitem[Shinn et~al.(2023)Shinn, Cassano, Gopinath, Narasimhan, and Yao]{shinn2023reflexion}
Shinn, N., Cassano, F., Gopinath, A., Narasimhan, K., and Yao, S.
\newblock Reflexion: Language agents with verbal reinforcement learning.
\newblock \emph{Advances in Neural Information Processing Systems}, 36:\penalty0 8634--8652, 2023.

\bibitem[Sun et~al.(2020)Sun, Wang, Liu, Miller, Efros, and Hardt]{sun2020test}
Sun, Y., Wang, X., Liu, Z., Miller, J., Efros, A., and Hardt, M.
\newblock Test-time training with self-supervision for generalization under distribution shifts.
\newblock In \emph{International conference on machine learning}, pp.\  9229--9248. PMLR, 2020.

\bibitem[Sun et~al.(2024)Sun, Li, Dalal, Xu, Vikram, Zhang, Dubois, Chen, Wang, Koyejo, et~al.]{sun2024learning}
Sun, Y., Li, X., Dalal, K., Xu, J., Vikram, A., Zhang, G., Dubois, Y., Chen, X., Wang, X., Koyejo, S., et~al.
\newblock Learning to (learn at test time): Rnns with expressive hidden states.
\newblock \emph{arXiv preprint arXiv:2407.04620}, 2024.

\bibitem[Sur{\'\i}s et~al.(2023)Sur{\'\i}s, Menon, and Vondrick]{suris2023vipergpt}
Sur{\'\i}s, D., Menon, S., and Vondrick, C.
\newblock Vipergpt: Visual inference via python execution for reasoning.
\newblock In \emph{Proceedings of the IEEE/CVF International Conference on Computer Vision}, pp.\  11888--11898, 2023.

\bibitem[Suzgun \& Kalai(2024)Suzgun and Kalai]{suzgun2024metaprompting}
Suzgun, M. and Kalai, A.~T.
\newblock Meta-prompting: Enhancing language models with task-agnostic scaffolding.
\newblock \emph{arXiv preprint arXiv:2401.12954}, 2024.

\bibitem[Suzgun et~al.(2019)Suzgun, Gehrmann, Belinkov, and Shieber]{suzgun2019memory}
Suzgun, M., Gehrmann, S., Belinkov, Y., and Shieber, S.~M.
\newblock Memory-augmented recurrent neural networks can learn generalized dyck languages.
\newblock \emph{arXiv preprint arXiv:1911.03329}, 2019.

\bibitem[Suzgun et~al.(2023{\natexlab{a}})Suzgun, Melas-Kyriazi, and Jurafsky]{suzgun2023follow}
Suzgun, M., Melas-Kyriazi, L., and Jurafsky, D.
\newblock Follow the wisdom of the crowd: Effective text generation via minimum bayes risk decoding.
\newblock In \emph{Findings of the Association for Computational Linguistics: ACL 2023}, pp.\  4265--4293, 2023{\natexlab{a}}.

\bibitem[Suzgun et~al.(2023{\natexlab{b}})Suzgun, Scales, Sch{\"a}rli, Gehrmann, Tay, Chung, Chowdhery, Le, Chi, Zhou, et~al.]{suzgun2023challenging}
Suzgun, M., Scales, N., Sch{\"a}rli, N., Gehrmann, S., Tay, Y., Chung, H.~W., Chowdhery, A., Le, Q., Chi, E., Zhou, D., et~al.
\newblock Challenging big-bench tasks and whether chain-of-thought can solve them.
\newblock In \emph{Findings of the Association for Computational Linguistics: ACL 2023}, pp.\  13003--13051, 2023{\natexlab{b}}.

\bibitem[Suzgun et~al.(2024)Suzgun, Shieber, and Jurafsky]{suzgun2024string2string}
Suzgun, M., Shieber, S.~M., and Jurafsky, D.
\newblock string2string: A modern python library for string-to-string algorithms.
\newblock In \emph{Proceedings of the 62nd Annual Meeting of the Association for Computational Linguistics (Volume 3: System Demonstrations)}, pp.\  278--285, 2024.

\bibitem[Syed et~al.(1999)Syed, Liu, and Sung]{syed1999handling}
Syed, N.~A., Liu, H., and Sung, K.~K.
\newblock Handling concept drifts in incremental learning with support vector machines.
\newblock In \emph{Proceedings of the fifth ACM SIGKDD international conference on Knowledge discovery and data mining}, pp.\  317--321, 1999.

\bibitem[Thrun \& Mitchell(1995)Thrun and Mitchell]{thrun1995lifelong}
Thrun, S. and Mitchell, T.~M.
\newblock Lifelong robot learning.
\newblock \emph{Robotics and autonomous systems}, 15\penalty0 (1-2):\penalty0 25--46, 1995.

\bibitem[Vu et~al.(2023)Vu, Iyyer, Wang, Constant, Wei, Wei, Tar, Sung, Zhou, Le, et~al.]{vu2023freshllms}
Vu, T., Iyyer, M., Wang, X., Constant, N., Wei, J., Wei, J., Tar, C., Sung, Y.-H., Zhou, D., Le, Q., et~al.
\newblock Freshllms: Refreshing large language models with search engine augmentation.
\newblock \emph{arXiv preprint arXiv:2310.03214}, 2023.

\bibitem[Wang et~al.(2020)Wang, Shelhamer, Liu, Olshausen, and Darrell]{wang2020tent}
Wang, D., Shelhamer, E., Liu, S., Olshausen, B., and Darrell, T.
\newblock Tent: Fully test-time adaptation by entropy minimization.
\newblock \emph{arXiv preprint arXiv:2006.10726}, 2020.

\bibitem[Wang et~al.(2023)Wang, Wei, Schuurmans, Le, Chi, Narang, Chowdhery, and Zhou]{wang2023selfconsistency}
Wang, X., Wei, J., Schuurmans, D., Le, Q.~V., Chi, E.~H., Narang, S., Chowdhery, A., and Zhou, D.
\newblock Self-consistency improves chain of thought reasoning in language models.
\newblock In \emph{The Eleventh International Conference on Learning Representations}, 2023.
\newblock URL \url{https://openreview.net/forum?id=1PL1NIMMrw}.

\bibitem[Wang et~al.(2024{\natexlab{a}})Wang, Gao, Chen, Jiang, Li, Yang, Yin, Li, Li, Yin, et~al.]{wang2024memoryllm}
Wang, Y., Gao, Y., Chen, X., Jiang, H., Li, S., Yang, J., Yin, Q., Li, Z., Li, X., Yin, B., et~al.
\newblock Memoryllm: Towards self-updatable large language models.
\newblock \emph{arXiv preprint arXiv:2402.04624}, 2024{\natexlab{a}}.

\bibitem[Wang et~al.(2024{\natexlab{b}})Wang, Ma, Zhang, Ni, Chandra, Guo, Ren, Arulraj, He, Jiang, Li, Ku, Wang, Zhuang, Fan, Yue, and Chen]{wang2024mmlupro}
Wang, Y., Ma, X., Zhang, G., Ni, Y., Chandra, A., Guo, S., Ren, W., Arulraj, A., He, X., Jiang, Z., Li, T., Ku, M., Wang, K., Zhuang, A., Fan, R., Yue, X., and Chen, W.
\newblock {MMLU}-pro: A more robust and challenging multi-task language understanding benchmark.
\newblock In \emph{The Thirty-eight Conference on Neural Information Processing Systems Datasets and Benchmarks Track}, 2024{\natexlab{b}}.
\newblock URL \url{https://openreview.net/forum?id=y10DM6R2r3}.

\bibitem[Wei et~al.(2022)Wei, Wang, Schuurmans, Bosma, Xia, Chi, Le, Zhou, et~al.]{wei2022chain}
Wei, J., Wang, X., Schuurmans, D., Bosma, M., Xia, F., Chi, E., Le, Q.~V., Zhou, D., et~al.
\newblock Chain-of-thought prompting elicits reasoning in large language models.
\newblock \emph{Advances in neural information processing systems}, 35:\penalty0 24824--24837, 2022.

\bibitem[Weston et~al.(2014)Weston, Chopra, and Bordes]{weston2014memory}
Weston, J., Chopra, S., and Bordes, A.
\newblock Memory networks.
\newblock \emph{arXiv preprint arXiv:1410.3916}, 2014.

\bibitem[Yang et~al.(2025)Yang, Yu, Zhang, Cao, Xu, Zhang, Gonzalez, and Cui]{yang2025buffer}
Yang, L., Yu, Z., Zhang, T., Cao, S., Xu, M., Zhang, W., Gonzalez, J.~E., and Cui, B.
\newblock Buffer of thoughts: Thought-augmented reasoning with large language models.
\newblock \emph{Advances in Neural Information Processing Systems}, 37:\penalty0 113519--113544, 2025.

\bibitem[Yao et~al.(2023)Yao, Yu, Zhao, Shafran, Griffiths, Cao, and Narasimhan]{yao2023tree}
Yao, S., Yu, D., Zhao, J., Shafran, I., Griffiths, T.~L., Cao, Y., and Narasimhan, K.
\newblock {Tree of Thoughts}: Deliberate problem solving with large language models, 2023.

\bibitem[Yuksekgonul et~al.(2025)Yuksekgonul, Bianchi, Boen, Liu, Lu, Huang, Guestrin, and Zou]{yuksekgonul2024textgrad}
Yuksekgonul, M., Bianchi, F., Boen, J., Liu, S., Lu, P., Huang, Z., Guestrin, C., and Zou, J.
\newblock Optimizing generative ai by backpropagating language model feedback.
\newblock \emph{Nature}, 639:\penalty0 609--616, 2025.

\bibitem[Zelikman et~al.(2022)Zelikman, Wu, Mu, and Goodman]{zelikman2022star}
Zelikman, E., Wu, Y., Mu, J., and Goodman, N.
\newblock Star: Bootstrapping reasoning with reasoning.
\newblock \emph{Advances in Neural Information Processing Systems}, 35:\penalty0 15476--15488, 2022.

\bibitem[Zhang et~al.(2024{\natexlab{a}})Zhang, Kang, Zhao, and Liu]{zhang-etal-2024-llm-based}
Zhang, K., Kang, Y., Zhao, F., and Liu, X.
\newblock {LLM}-based medical assistant personalization with short- and long-term memory coordination.
\newblock In Duh, K., Gomez, H., and Bethard, S. (eds.), \emph{Proceedings of the 2024 Conference of the North American Chapter of the Association for Computational Linguistics: Human Language Technologies (Volume 1: Long Papers)}, pp.\  2386--2398, Mexico City, Mexico, June 2024{\natexlab{a}}. Association for Computational Linguistics.
\newblock \doi{10.18653/v1/2024.naacl-long.132}.
\newblock URL \url{https://aclanthology.org/2024.naacl-long.132/}.

\bibitem[Zhang et~al.(2022)Zhang, Levine, and Finn]{zhang2022memo}
Zhang, M., Levine, S., and Finn, C.
\newblock Memo: Test time robustness via adaptation and augmentation.
\newblock \emph{Advances in neural information processing systems}, 35:\penalty0 38629--38642, 2022.

\bibitem[Zhang et~al.(2024{\natexlab{b}})Zhang, Patil, Jain, Shen, Zaharia, Stoica, and Gonzalez]{zhang2024raft}
Zhang, T., Patil, S.~G., Jain, N., Shen, S., Zaharia, M., Stoica, I., and Gonzalez, J.~E.
\newblock {RAFT}: Adapting language model to domain specific {RAG}.
\newblock In \emph{First Conference on Language Modeling}, 2024{\natexlab{b}}.
\newblock URL \url{https://openreview.net/forum?id=rzQGHXNReU}.

\bibitem[Zhong et~al.(2022)Zhong, Lei, and Chen]{zhong-etal-2022-training}
Zhong, Z., Lei, T., and Chen, D.
\newblock Training language models with memory augmentation.
\newblock In Goldberg, Y., Kozareva, Z., and Zhang, Y. (eds.), \emph{Proceedings of the 2022 Conference on Empirical Methods in Natural Language Processing}, pp.\  5657--5673, Abu Dhabi, United Arab Emirates, December 2022. Association for Computational Linguistics.
\newblock \doi{10.18653/v1/2022.emnlp-main.382}.
\newblock URL \url{https://aclanthology.org/2022.emnlp-main.382/}.

\bibitem[Zhou et~al.(2022)Zhou, Sch{\"a}rli, Hou, Wei, Scales, Wang, Schuurmans, Cui, Bousquet, Le, et~al.]{zhou2022least}
Zhou, D., Sch{\"a}rli, N., Hou, L., Wei, J., Scales, N., Wang, X., Schuurmans, D., Cui, C., Bousquet, O., Le, Q., et~al.
\newblock Least-to-most prompting enables complex reasoning in large language models.
\newblock \emph{arXiv preprint arXiv:2205.10625}, 2022.

\end{thebibliography}
\bibliographystyle{icml2025}

\newpage
\appendix

\clearpage
\section{Background \& Related Work}
\label{sec:background}

\subsection{Test-time learning (online learning)}

Test-time learning---also referred to as online or incremental learning (adaptation)---encompasses a family of methods in which a stochastic model updates its predictions by incorporating information seen during inference, without undergoing conventional, full-scale offline finetuning. Early versions of test-time adaptation focused on local or transductive learning, where a model re-fit or re-weighted its parameters with each new test instance or batch~\citep[][\emph{inter alia}]{mccloskey1989catastrophic,thrun1995lifelong,amari1998natural,syed1999handling,bottou2003large,bottou2005online}. In computer vision, for example, methods like test-time training have been shown to mitigate domain shifts by optimizing a self-supervised loss on incoming data~\citep{wang2020tent,sun2020test,liu2021tttplus,boudiaf2022parameter,niu2022efficient,zhang2022memo,sun2024learning}. In the context of natural-language generation, test-time adaptation has appeared under terms such as ``dynamic evaluation''~\citep{mikolov2010recurrent,graves2013generating,krause2019dynamic,rannen2024revisiting}, in which a language model is updated with gradient steps on the test-time data itself.

However, directly updating language model weights at test time can be computationally expensive and requires the capacity to modify parameters. For large-scale, black-box APIs (e.g., GPT-3 or Claude), one often lacks the ability to perform parameter updates easily, thereby making such an approach difficult, if not completely infeasible~\citep{shietal2024replug}. To address this, a growing body of work has explored parameter-free adaptation, whereby one structurally modifies immediate model inputs (e.g., prompting) or draws from external memory to “update” the model’s effective reasoning. Our approach aligns with this direction by allowing an LM to iteratively record solutions, explanations, or heuristics in an external memory component over successive interactions, avoiding weight updates entirely. 

In the broader test-time learning literature, reflexive, compositional, and iterative refinement approaches like Reflexion~\citep{shinn2023reflexion}, Self-Refine~\citep{madaan2023self}, (Self-)Critic~\citep{gou2023critic}, Chameleon~\citep{lu2023chameleon}, Meta-Prompting~\citep{suzgun2024metaprompting}, and Self-RAG~\citep{asai2023self} \emph{inter alia}, use feedback loops or verification mechanisms to correct mistakes in solutions. TextGrad~\citep{yuksekgonul2024textgrad} similarly draws on the notion of ``textual gradients'' as an alternative to parameter-based gradients and provides a pathway for improvement based on the content of mistakes. Our proposed DC framework differs by focusing explicitly on storing generalizable heuristics, solutions, or meta-level insights that can be repeatedly retrieved and applied across tasks, not just to correct a single solution. Furthermore, DC does not require a new training loop for each batch or scenario; instead, the memory itself is updated to reflect newly found solutions, errors, or strategies without touching the model weights.

\subsection{Test-time compute and reasoning}

It is now widely known and accepted that contemporary LLMs such as GPT-4 can exhibit substantial improvements in reasoning and generation capability when additional compute is devoted to inference-time strategies (e.g., chain-of-thought prompting~\citep{wei2022chain,kojima2022large,zhou2022least}, tree-of-thought expansions~\citep{yao2023tree,long2023large}, minimum Bayes risk decoding~\citep{suzgun2023follow,shi2022natural,golovneva2023pathfinder}, majority-vote sampling~\citep{wang2023selfconsistency}). Prompting methods such as Tree-of-Thought~\citep{yao2023tree}, Graph-of-Thought~\citep{besta2024graph}, and other non-linear compositional reasoning paradigms systematically enlarge the inference-time search space. They allow models to explore various reasoning paths and exploit consensus or iterative corrections to arrive at more accurate and reliable conclusions~\citep{wei2022chain,wang2023selfconsistency}.

However, these expansions come at the cost of increased computational overhead per test instance~\citep{yao2023tree}. They are, however, typically ephemeral: once a solution is generated, subsequent tasks or input samples do not generally benefit from the heavy compute spent earlier, unless the user manually engineers advanced prompt-sharing or in-context demonstration strategies. \emph{Cf.}~\citep{zelikman2022star}. Our work, on the other hand, aims to reduce repeated overhead across multiple test instances of a similar domain by building a memory that persists from one query to the next. This memory not only reduces repetitive mistakes, but also consolidates and codifies robust solution strategies—effectively amortizing or “sharing” the cost of initial reflection across future tasks.\footnote{Some lines of work---such as majority voting or sampling-based self-consistency---combine multiple inference passes for a single question but still lack a persistent knowledge base that spans different queries. DC differs in that we treat consecutive tasks in a sequence as a chance to refine a persistent, external store of learned lessons. The memory curation step selectively compiles relevant solutions, heuristics, expansions, or code blocks into a form that can be reused for upcoming queries. Thus, while the compute for the first few tasks may be higher, future tasks become simpler because the system can consult and adapt previously curated knowledge. This approach echoes the underlying motivation of test-time training---performing ongoing improvement at inference---but capitalizes on a cheap, external memory update in lieu of repeated or expensive parameter updates.}

Another related thread involves tool usage or code execution~\citep{schick2023toolformer,lu2023chameleon,shen2023hugginggpt,qin2023toolllm,suris2023vipergpt,suzgun2024metaprompting}. These studies have explored how LLMs can call external Python interpreters, symbolic solvers, or other specialized services and APIs to offload complex computations. Our empirical findings too illustrate that once an LLM under DC recognizes a systematic way (e.g., Python-based brute force algoritgm) to handle a certain class of problems (like arithmetic puzzles), it can store that approach in memory and repeatedly retrieve it. Thus, DC not only invests extra compute in a single session but spreads that computational benefit across multiple interactions, effectively learning to use tools more consistently and reliably over time.

\subsection{Memory-augmented generation and reasoning}

Augmenting language models with external memory has seen renewed interest in recent years~\citep{munkhdalai2019metalearned,guu2020retrieval,Khandelwal2020Generalization,bulatov2022recurrent,borgeaud2022improving,zhong-etal-2022-training,feng2022learn,he2024camelot,wang2024memoryllm}---\emph{see also}~\citep{graves2014neural,weston2014memory,joulin2015inferring,suzgun2019memory} for early studies. Modern retrieval-augmented LLM approaches generally consult an external corpus of documents (i.e., a knowledge base) to improve factuality and reduce hallucination~\citep{lewis2020retrieval,lazaridou2023internetaugmented,vu2023freshllms,zhang2024raft}, but the retrieval corpus is almost always fixed prior to inference and does not evolve over time. These methods have been especially effective for open-domain question answering \cite{lewis2020retrieval,guu2020retrieval,karpukhin2020dense}, where the model’s own parameters may not hold all relevant knowledge. In practice, retrieval augmentation typically involves selecting and concatenating top-$k$ passages from a knowledge-base---while useful for factual queries, the approach, however, does not inherently solve iterative improvement or learning from mistakes in the sense of building upon prior solutions at inference time.

Another line of research more closely aligns with our vision by storing not just reference knowledge but also the reasoning processes and solution strategies of language models. Several recent works have explored this direction. Thought-Retriever~\citep{feng2024thoughtretriever} logs the model’s chain-of-thought from past queries and uses them for new, analogous queries. Buffer-of-Thoughts~\citep[BoT;][]{yang2025buffer} takes a slightly different approach by distilling high-level ``thought templates'' from problem-solving processes, though it relies on predefined templates that seem to be tailored towards specific task types that were considered in their experiments. \citet{madaan2022memory} have demonstrated that deployed models like GPT-3 can be improved through memory mechanisms that capture user feedback on errors, preventing similar mistakes in future interactions. \citet{zhang-etal-2024-llm-based} have proposed a dual memory architecture combining long-term and short-term storage for medical applications, though their approach requires fine-tuning to incorporate new knowledge.

While these works reveal the many strategies for harnessing memory or feedback, DC emphasizes selectively storing the most relevant insights and heuristics. DC aims to avoid naive accumulation of full raw transcripts and ephemeral chain-of-thought expansions that can lead to memory bloat. Moreover, unlike methods that assume the model can be retrained or finetuned to incorporate memory items, we remain fully external and training-free; this aligns with ``plug-and-play'' usage principle, in which an off-the-shelf model is augmented by an external memory that it reads from and writes to, but does not require any gradient-based adaptation.

\onecolumn

\clearpage

\section{Additional Figures and Tables}

\subsection{Performance Comparison of Baseline and DC-RS Approaches}

\begin{figure*}[!h]
\centering
\includegraphics[width=0.55\textwidth]{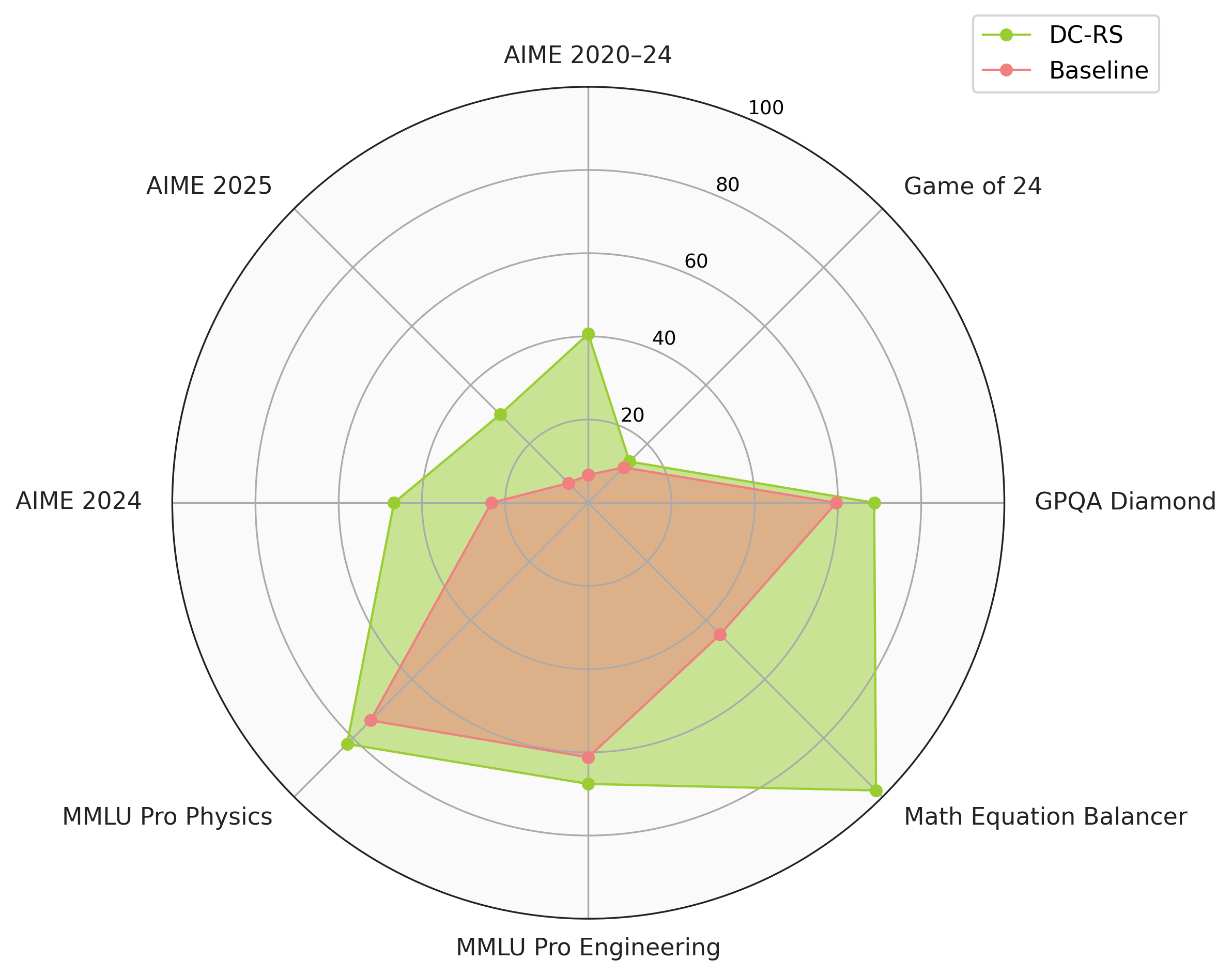}
\caption{
\textbf{Overall performance of Claude 3.5 Sonnet} under the baseline prompting approach with minimal instructions (Baseline) and Dynamic Cheatsheet with Retrieval \& Synthesis (DC-RS). }
\label{fig:claude-radar-chart}
\end{figure*}

\begin{figure*}[!h]
\centering
\includegraphics[width=0.55\textwidth]{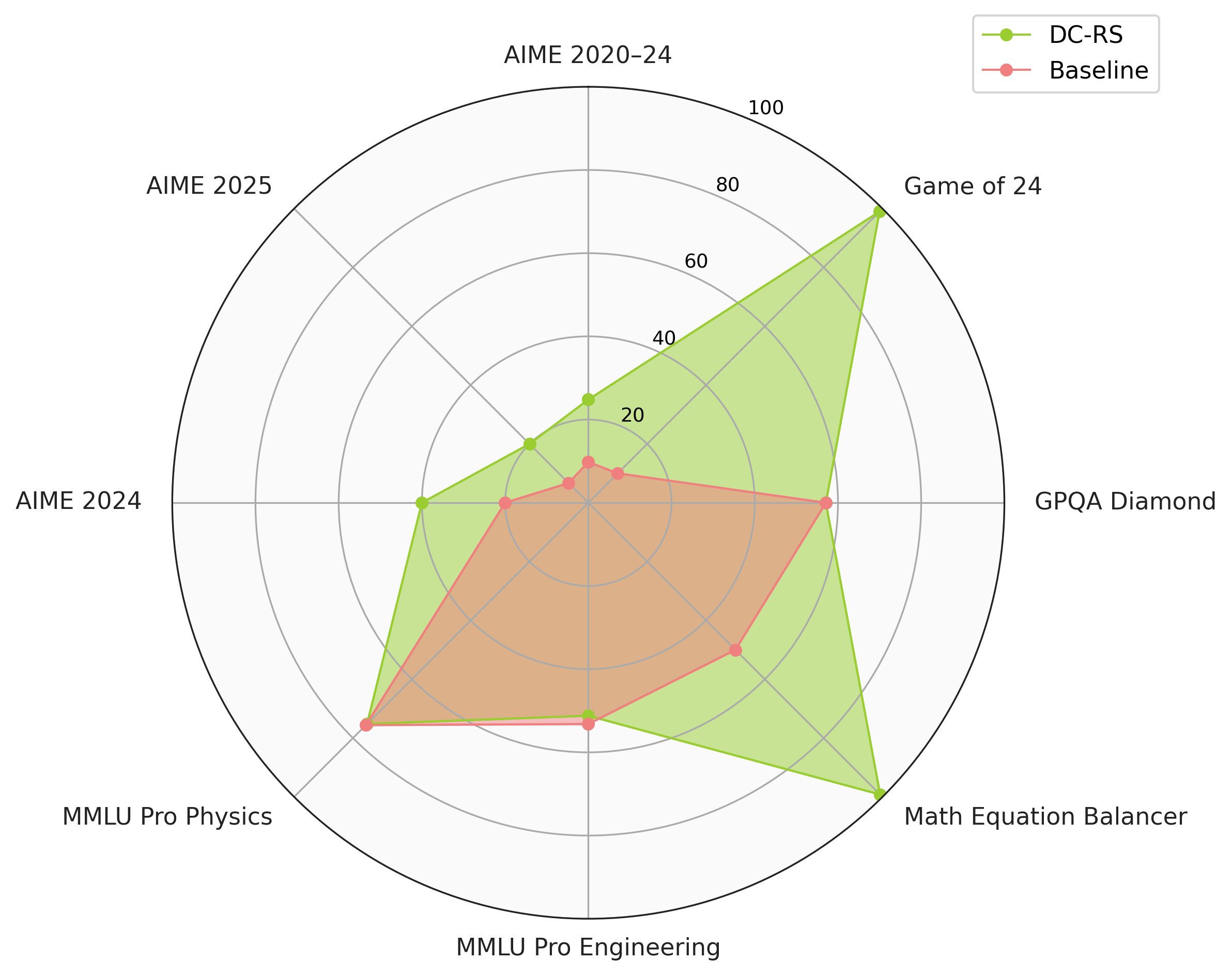}
\caption{
\textbf{Overall performance of GPT-4o} under the baseline prompting approach with minimal instructions (Baseline) and Dynamic Cheatsheet with Retrieval \& Synthesis (DC-RS). }
\label{fig:gpt4o-radar-chart}
\end{figure*}

\clearpage

\subsection{Clustering of Errors and Corrections}

\begin{figure}[h]
\centering
\includegraphics[width=0.99\textwidth]{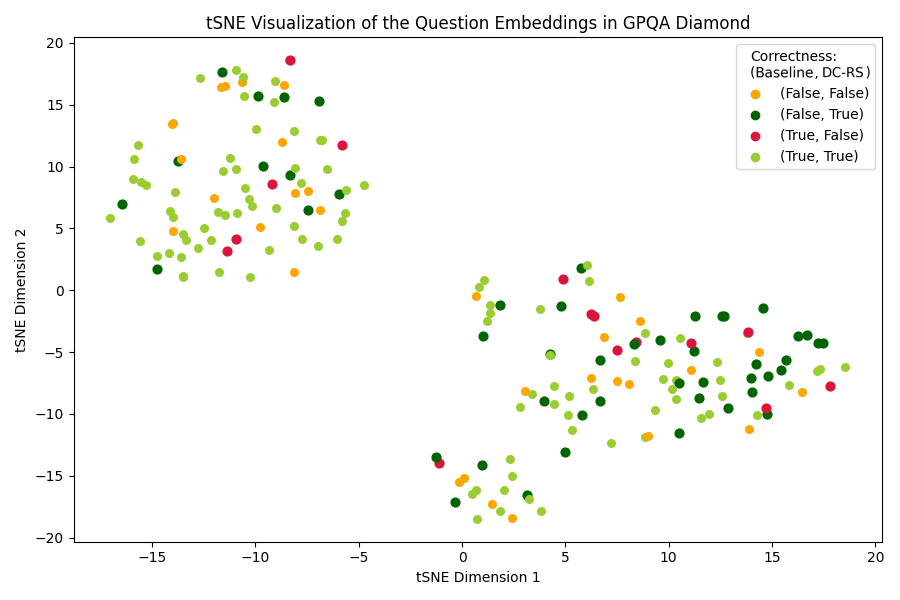}
\caption{
t-SNE visualization of the embeddings of the raw questions in GPQA-Diamond. Note that correct and incorrect answers often cluster in latent embedding space. DC can help transfer learned strategies within these clusters, but without careful curation, erroneous heuristics may also spread, thus requiring careful memory refinement and verification of solution strategies. 
}
\label{fig:question-embeddings}
\end{figure}

\clearpage

\subsection{Evolution of Memory Content under Dynamic Cheatsheet}

\begin{figure}[h]
\centering
\includegraphics[width=0.99\linewidth]{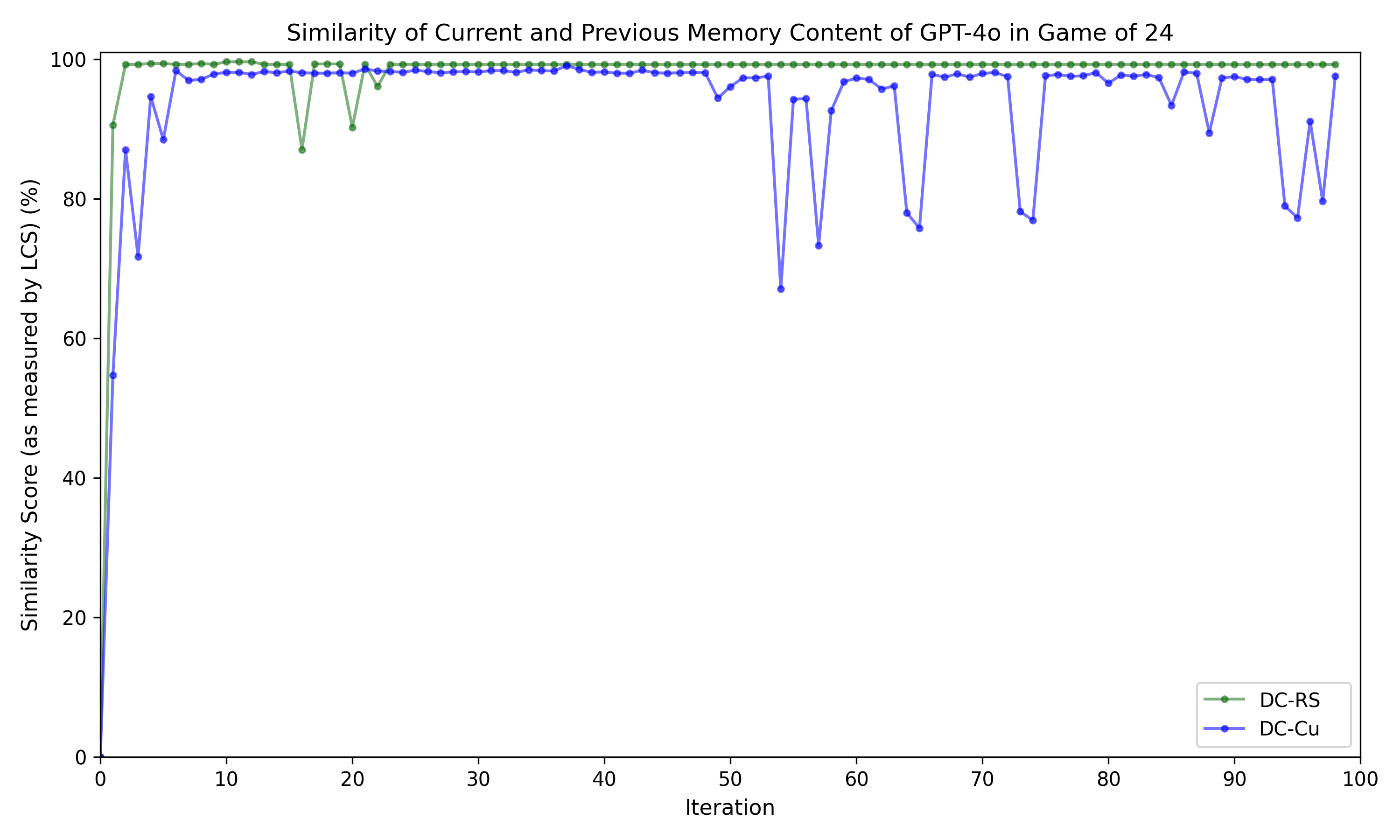}
\caption{
This figure illustrates how memory content of GPT-4o evolves over time in Game of 24, quantified using a longest-common-subsequence (LCS)-similarity metric~\citep{suzgun2024string2string} between consecutive states (measured at the word level). While both DC-Cu and DC-RS show high stability after the first few iterations, DC-Cu experiences slightly greater fluctuations in the second half of inference. 
}
\label{fig:memory-evolution}
\end{figure}

\clearpage

\subsection{Solution Generator and Memory Curator Prompts}

\subsubsection{Prompt Used by the Generator Model in Baseline}
\begin{figure}[h]
\centering
\includegraphics[width=0.85\textwidth]{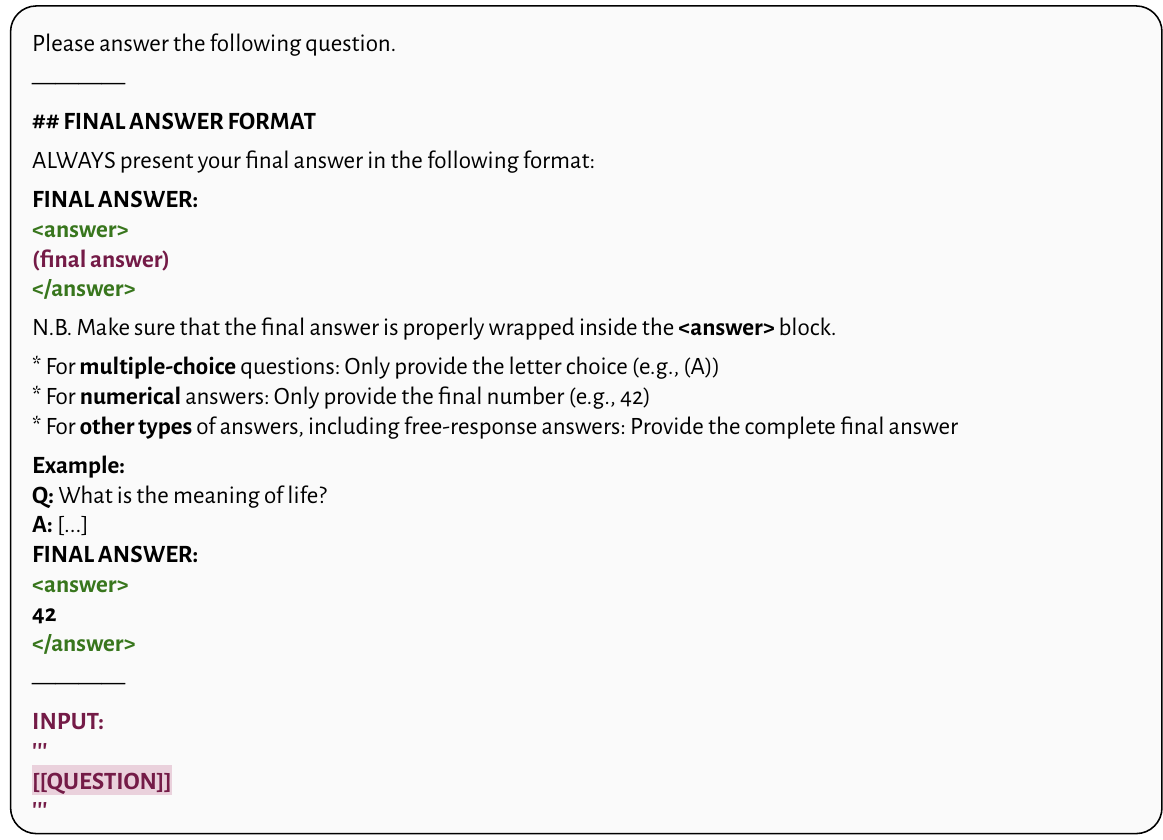}
\caption{
Prompt used in the baseline (BL) approach, where the model receives minimal instructions. The prompt simply asks the model to answer the given question without any structured guidance, additional reasoning steps, or tool-use encouragement. This setup represents a traditional one-off inference method, reflecting how LLMs typically operate by default.
}
\label{fig:baseline-prompt}
\end{figure}

\clearpage

\subsubsection{Prompt Used by the Generator Model in DR, FH, and DC Approaches}

\begin{figure}[!h]
\centering
\includegraphics[width=0.60\textwidth]{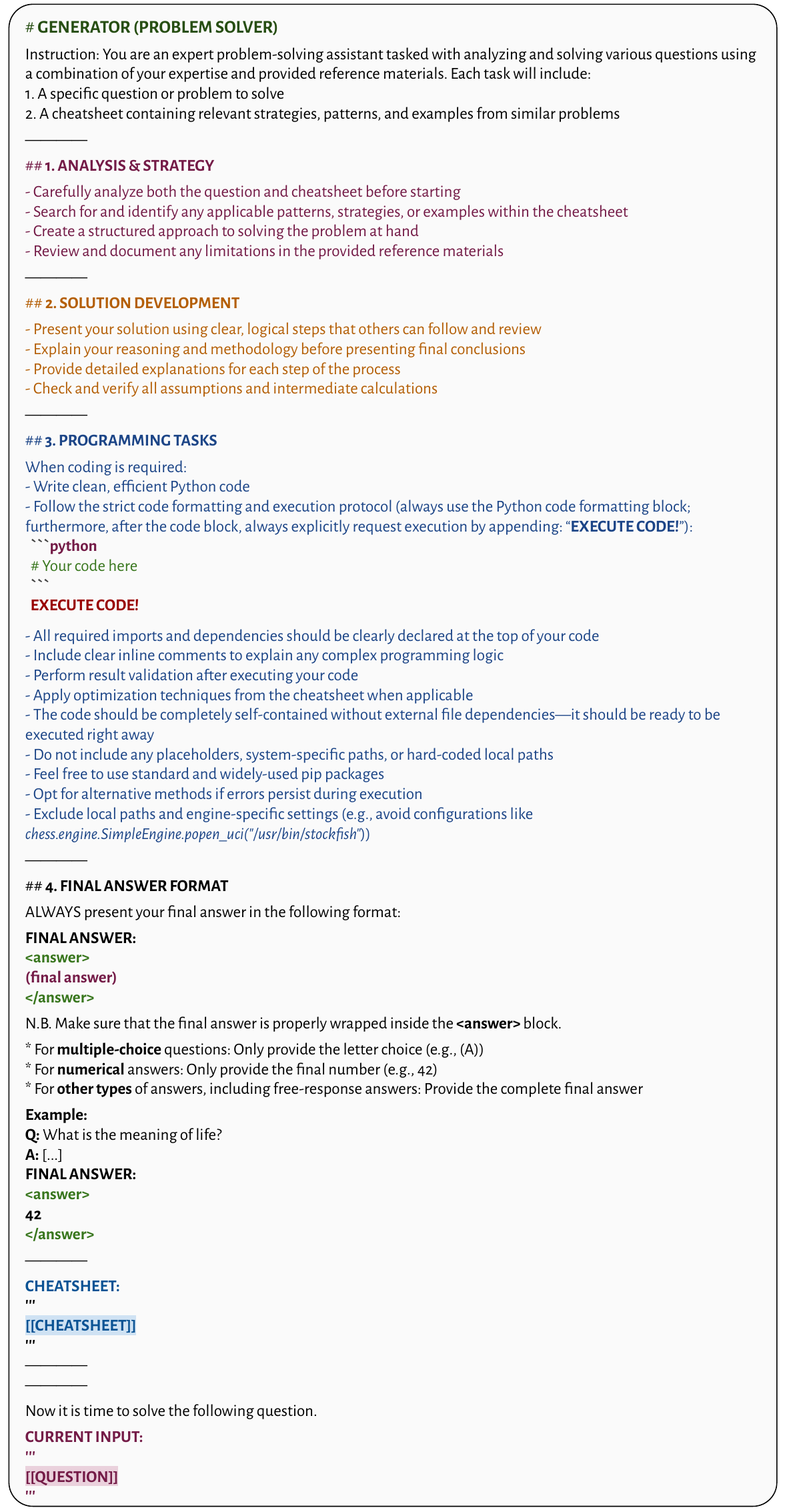}
\caption{
Generator prompt used in the DR, FH, and DC approaches, where the model receives structured high-level instructions on solution development, strategy selection, and tool usage. This prompt explicitly encourages Python code generation and execution for computational tasks. Notably, this same structured prompt is used in all non-BL methods, including DC-$\emptyset$, DR, FH, DC-Cu, and DC-RS. We also remark that during the initial phases of our experiments, we used ``cheatsheet'' and ``memory'' interchangeably to describe stored problem-solving content. However, to maintain consistency, we formally define $M_i$ as memory throughout this paper. Since this was purely a semantic choice, we did not find it necessary to rerun our experiments to reflect this terminology shift.
}
\label{fig:generator-prompt}
\end{figure}

\clearpage

\subsubsection{Prompt Used by the Memory Curation Model under DC-RS}

\begin{figure*}[!h]
\centering
\includegraphics[width=0.65\textwidth]{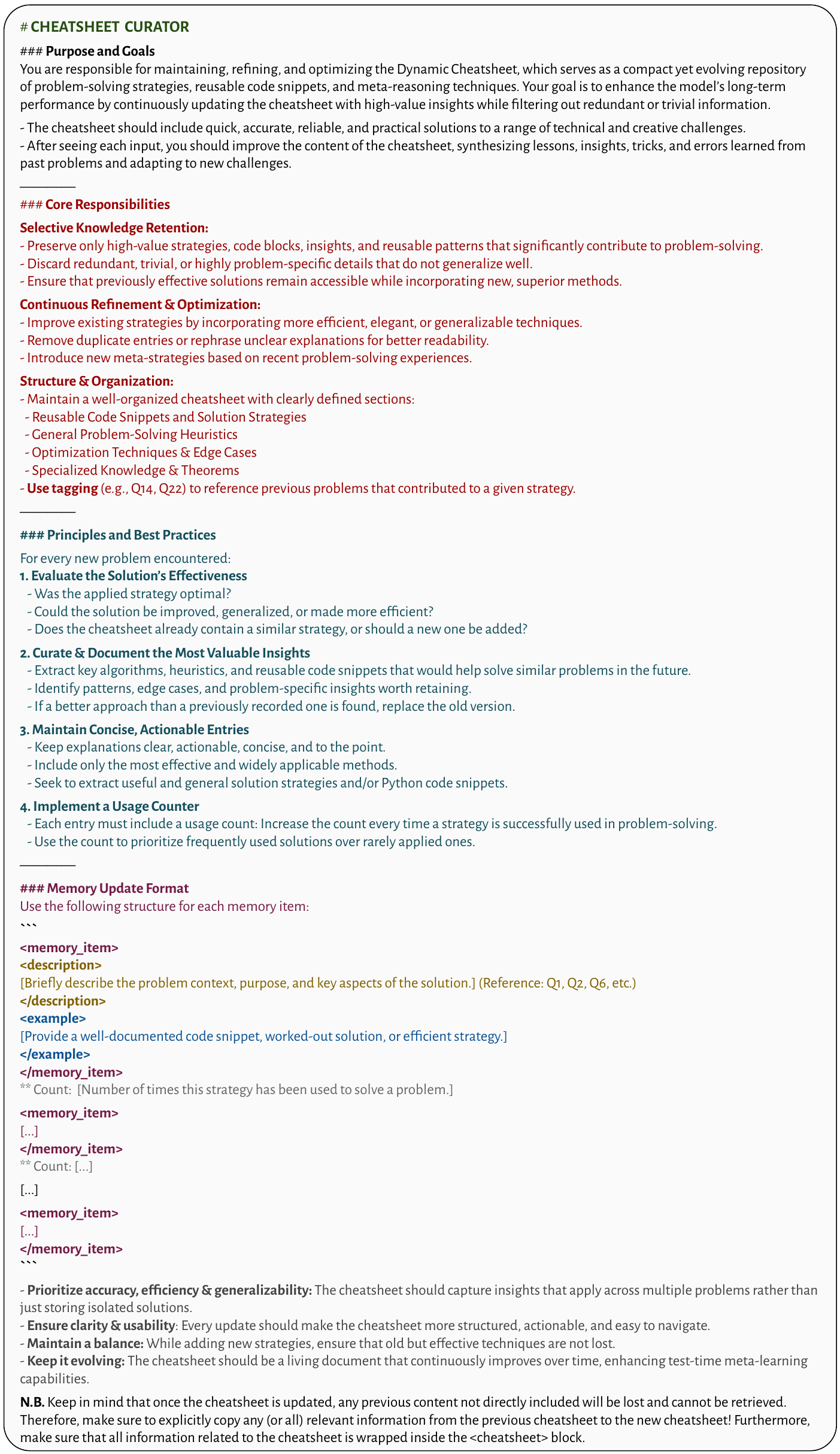}
\caption{
Prompt used for the memory curator under DC-RS, which is responsible for maintaining an evolving repository of problem-solving strategies, code snippets, and heuristics. The curator selectively retains high-value insights, refines existing strategies, and organizes memory efficiently. This ensures the memory (cheatsheet) remains concise, generalizable, and action-oriented, continuously improving test-time reasoning. (Once again, we note that during the initial phases of our experiments, we used ``cheatsheet'' and ``memory'' interchangeably to describe stored problem-solving content. However, to maintain consistency, we formally define $M_i$ as memory throughout this paper. Since this was purely a semantic choice, we did not find it necessary to rerun our experiments to reflect this terminology shift.)
}
\label{fig:curator-prompt-rs}
\end{figure*}

\clearpage

\begin{figure*}[h]
\centering
\includegraphics[width=0.65\textwidth]{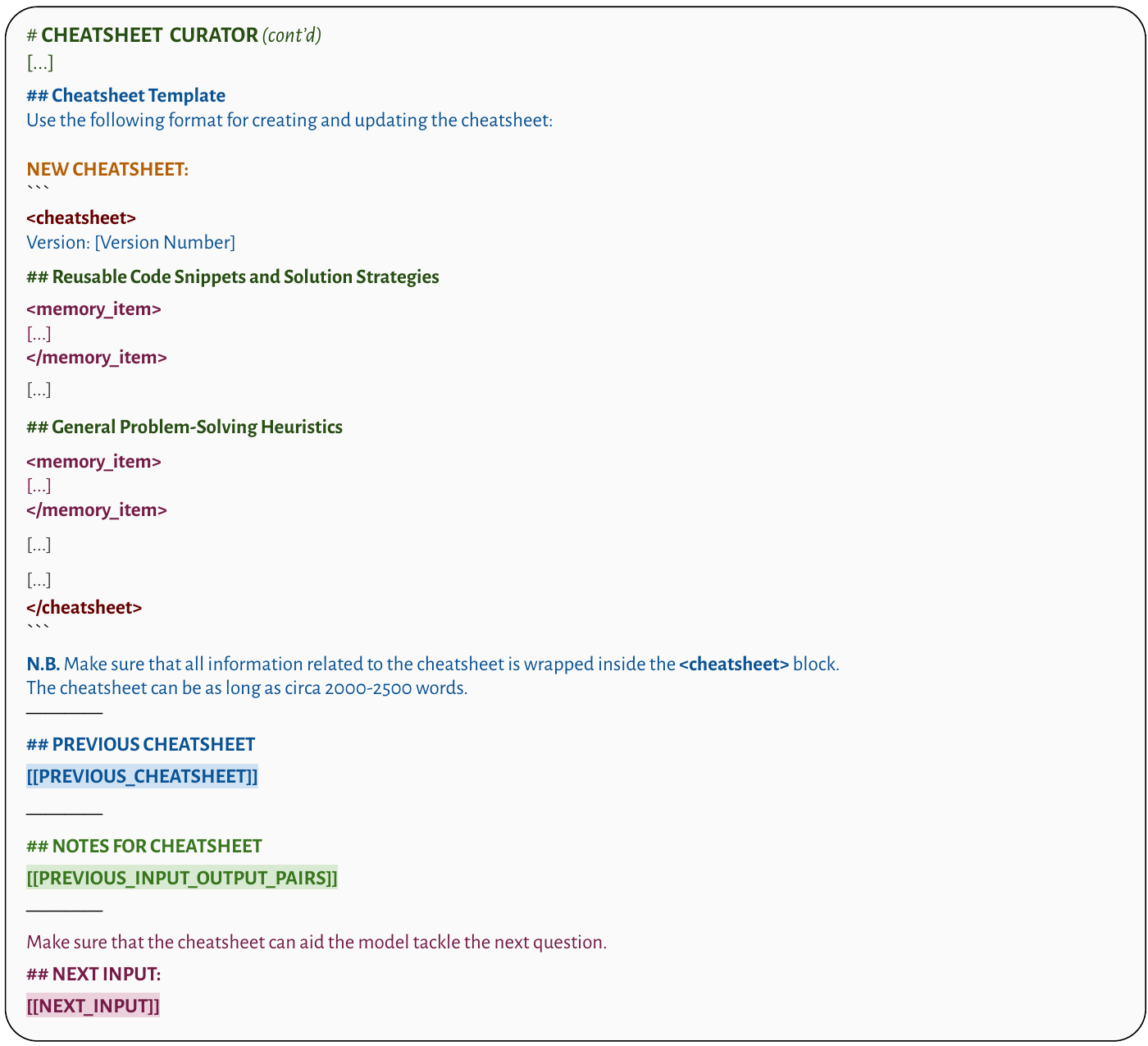}
\caption{
The rest of the prompt used by the memory curator under DC-RS (Figure~\ref{fig:curator-prompt-rs}).
}
\label{fig:curator-prompt-rs-rest}
\end{figure*}

\end{document}